\pdfoutput=1

\documentclass[11pt]{article}

\usepackage{emnlp2021}

\usepackage{afterpage}
\usepackage{amssymb}
\usepackage{amsfonts}
\usepackage[russian, english]{babel}
\usepackage{CJKutf8}
\usepackage{examples}
\usepackage{eurosym}
\usepackage[T1]{fontenc}
\usepackage{graphicx}
\usepackage[utf8]{inputenc}
\usepackage{latexsym}
\usepackage{longtable}
\usepackage{microtype}
\usepackage{multicol,lipsum}
\usepackage{multirow}
\usepackage{pifont}
\usepackage{pdflscape}
\usepackage{subfiles}
\usepackage{times}
\usepackage{url}
\usepackage{xcolor}

\title{English Machine Reading Comprehension Datasets: A Survey}

\author{
  Daria Dzendzik \\
  ADAPT Centre \\
  Dublin City University \\ 
  Dublin, Ireland \\
  \tt daria.dzendzik\\\tt @adaptcentre.ie
  \\\And
  Carl Vogel \\
  School of Computer\\ 
  Science and Statistics\\
  Trinity College Dublin\\
  the University of Dublin\\
  Dublin, Ireland \\ 
  \tt vogel@tcd.ie
  \\\And Jennifer Foster\\
  School of Computing  \\
  Dublin City University \\ 
  Dublin, Ireland\\ 
  \tt jennifer.foster\\\tt @dcu.ie\\
  }

\begin{document}
\maketitle
\begin{abstract}
This paper surveys 60 English Machine Reading Comprehension datasets, with a view to providing a convenient resource for other researchers interested in this problem. 
We categorize the datasets according to their question and answer form and compare them across various dimensions including size, vocabulary, data source, method of creation, human performance level, and first question word. 
Our analysis reveals that Wikipedia is by far the most common data source and that there is a relative lack of \textit{why}, \textit{when}, and \textit{where} questions across datasets. 
\end{abstract}

\section{Introduction}

\begin{figure}[ht]
    \includegraphics[width=7.5cm]{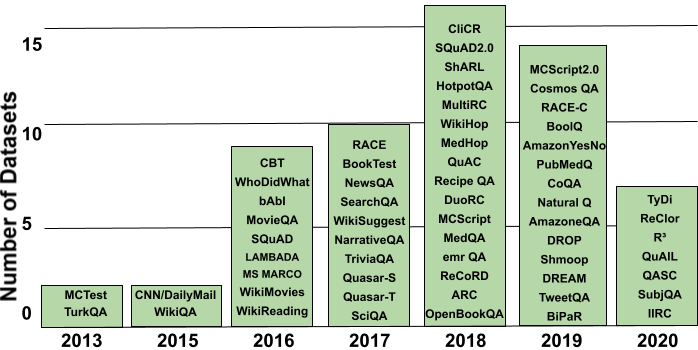}
    \caption{English MRC datasets released per year}
    \label{pic:years}
 \end{figure}

Reading comprehension is often tested by measuring a person or system’s ability to answer questions on a given text. Machine reading comprehension (MRC) datasets have proliferated in recent years, particularly for the English language -- see Fig.~\ref{pic:years}. The aim of this paper is to make sense of this landscape by providing as extensive as possible a survey of English MRC datasets.
The survey has been carried out with the  following audience in mind: (1) those who are new to the field and would like to get a concise yet informative overview of English MRC datasets;
(2) those who are planning to create a new MRC dataset;
(3) MRC system developers, interested in designing the appropriate architecture for a particular dataset, choosing appropriate datasets for a particular architecture, or finding compatible datasets for use in transfer or joint learning.

Our survey takes a mostly structured form, with the following information presented for each dataset: size, data source, creation method, human performance level, whether the dataset has been ``solved’’, availability of a leaderboard, the most frequent first question token, and whether the dataset is publicly available. 
We also categorise each dataset by its question/answer type. 
 
 Our study contributes to the field as follows:
 \begin{enumerate}
  \setlength{\itemsep}{-2.5pt}
     \item it describes and teases apart the ways in which MRC datasets can vary according to their question and answer types;
     \item it provides analysis in table and figure format to facilitate easy comparison between datasets;
     \item by providing a systematic comparison, and by reporting the ``solvedness'' status of a dataset, it brings the attention of the community to less popular and relatively understudied datasets;
     \item it contains per-dataset statistics such as number of instances, average question/passage/answer length, vocabulary size and text domain which can be used to estimate the computational requirements for training an MRC system. 
 \end{enumerate}

\section{Question, Answer, and Passage Types}\label{sec:types}
\begin{figure*}[ht]
\centering
     \includegraphics[width=12cm]{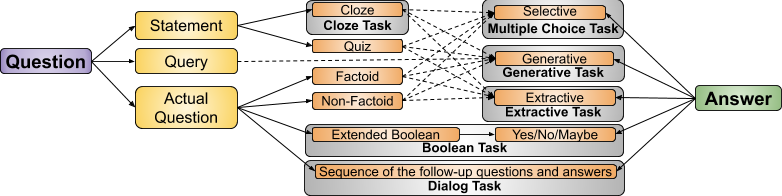}
     \caption{Hierarchy of types of question and answer and the relationships between them.
     $\rightarrow$ indicates a subtype whereas $\dashrightarrow$  indicates inclusion. 
     }\label{pic:qa_relation}
 \end{figure*}

All MRC datasets  in this survey have three components: \textit{passage}, \textit{question}, and \textit{answer}.\footnote{We briefly describe datasets which do not meet this criteria in  Appendix~\ref{sec:other}  and explain why we exclude them.} We begin with a categorisation based on the types of answers and the way the question is formulated.
We divide questions into three main categories:
\textit{Statement}, \textit{Query}, and \textit{Question}. Answers are divided into the following categories:
\textit{Cloze}, \textit{Multiple Choice}, \textit{Boolean}, \textit{Extractive}, \textit{Generative}.
The relationships between question and answer types are illustrated in Fig.~\ref{pic:qa_relation}.
In what follows, we briefly describe each question and answer category, followed by a discussion of passage types, and dialog-based datasets.
\subsection{Answer Type} \label{subsec:answertype}
    \paragraph{Cloze}The question is formulated as a sentence with a missing word or phrase which should be inserted into the sentence or should complete the sentence.
    The answer candidates may be included as in (\ref{ex:close1}) from ReciteQA~\cite{yagcioglu-etal-2018-recipeqa}, and may not, as in (\ref{ex:close2}) from CliCR~\cite{suster-daelemans-2018-clicr}.
    \begin{examples}
        \setlength{\itemsep}{-2.5pt}
       \item \textbf{Passage (P):} 
       \textit{You will need 3/4 cup of  blackberries ... Pour the mixture into cups and insert a popsicle stick in it or pour it in a popsicle maker. Place the cup ...
       in the freezer. ...}
\\ 
        \textbf{Question (Q): } \textit{Choose the best title for the missing blank to correctly complete the recipe.} 
        \textit{Ingredients, 
       \_, 
       Freeze,
        Enjoying} \\ 
        \textbf{Candidates (AC):} 
        \textit{(A) Cereal Milk Ice Cream
        (B) Ingredients
        (C) Pouring
        (D) Oven} \\ \textbf{Answer (A):} C
        \label{ex:close1}
        
        \item \textbf{P:} \textit{
        ... intestinal perforation in dengue is very rare and has been reported only in eight patients 
        ...}
        \textbf{Q:} \textit{Perforation peritonitis is a \_.}
        \textbf{Possible A:} \textit{very rare complication of dengue} 
        \label{ex:close2}

\end{examples}

    \paragraph{Selective or Multiple Choice (MC)}A number of options is given for each question, and the correct one(s) 
    should be selected, e.g.~(\ref{ex:mc}) from MCTest \cite{richardson2013mctest}. 
    
    \begin{examples}
          \setlength{\itemsep}{0pt}
        \item \textbf{P:} \textit{It was Jessie Bear's birthday. She ...}
        \textbf{Q:} \textit{Who was having a birthday?}
        \textbf{AC:} \textit{(A) Jessie Bear
        (B) no one
        (C) Lion
        (D) Tiger} 
        \textbf{A:} A
        \label{ex:mc}
    \end{examples}
    We distinguish cloze multiple choice datasets from other multiple choice datasets. The difference is the form of question: in the cloze datasets, the answer is a missing part of the question context and, combined together, they form a grammatically correct sentence, whereas for other multiple choice datasets, the question has no missing words.

    \paragraph{Boolean}A Yes/No answer is expected, e.g.~(\ref{ex:bool1}) from the BoolQ dataset~\cite{clark-etal-2019-boolq}.
    Some datasets which we include here have a third  \textit{``Cannot be answered"} or \textit{``Maybe"} option, e.g. (\ref{ex:bool3}) from PubMedQuestions~\cite{pubmedqa}. 
    \begin{examples}
      \setlength{\itemsep}{0pt}
            \item \textbf{P:} \textit{The series is filmed partially in Prince Edward Island as well as locations in ...
            }
            \textbf{Q:} \textit{Is anne with an e filmed on pei?} 
            \textbf{A:} \textit{Yes} 
            \label{ex:bool1}

            \item \textbf{P:} \textit{ ... Young adults whose families were abstainers in 2000 drank substantially less across quintiles in 2010 than offspring of non-abstaining families. The difference, however, was not statistically significant between quintiles of the conditional distribution. Actual drinking levels in drinking families were not at all or weakly associated with drinking in offspring. ...} 
            \textbf{Q:} \textit{Does the familial transmission of drinking patterns persist into young adulthood?} 
            \textbf{A:} \textit{Maybe} 
            \label{ex:bool3}
    \end{examples}
    \paragraph{Extractive or Span Extractive}The answer is a substring of the passage. In other words, the task is to determine the answer character start and end index  in the original passage, 
    as shown in (\ref{ex:extractive}) from SQuAD~\cite{rajpurkar-etal-2016-squad}.
    
    \begin{examples}
      \setlength{\itemsep}{0pt}
            \item \textbf{P:} \textit{With Rivera having been a linebacker with the Chicago Bears in Super Bowl XX, ....
            }
            \textbf{Q:} \textit{What team did Rivera play for in Super Bowl XX?}
            \textbf{A:} \textit{46-59: Chicago Bears} 
            \label{ex:extractive}
    \end{examples}

    \paragraph{Generative or Free Form Answer}The answer must be generated based on information presented in the passage. Although the answer might be in the text, as illustrated in (\ref{ex:generative}) from NarrativeQA~\cite{kocisky-etal-2018-narrativeqa}, no passage index connections are provided. 
    
    \begin{examples}
      \setlength{\itemsep}{0pt}
            \item \textbf{P:} \textit{...Mark decides to broadcast his final message as himself. They finally drive up to the crowd of protesting students, ....
            The police step in and arrest Mark and Nora....}
            \textbf{Q:} \textit{What are the students doing when Mark and Nora drive up?}
            \textbf{A:} \textit{Protesting.} 
            \label{ex:generative}
    \end{examples}
    
\subsection{Question Type}

    \paragraph{Statement}The question is a declarative sentence and used in cloze, e.g.~(\ref{ex:close1}-\ref{ex:close2}), and quiz questions, e.g.~(\ref{ex:quiz}) from SearchQA~\cite{DBLP:journals/corr/DunnSHGCC17}. 
    
    \begin{examples}
      \setlength{\itemsep}{0pt}
            \item 
            \textbf{P:} \textit{Jumbuck (noun) is an Australian English term for sheep, 
            ...}
            \textbf{Q:} \textit{Australians call this animal a jumbuck or a monkey}
            \textbf{A:} \textit{Sheep} 
            \label{ex:quiz}
    \end{examples}

    \paragraph{Question}is an actual question in the standard sense of the word, e.g. (\ref{ex:mc})-(\ref{ex:generative}). 
    Usually
    questions are divided into Factoid (\textit{Who? Where? What? When?}), Non-Factoid (\textit{How? Why?}), and Yes/No. 

    \paragraph{Query}The question is formulated to obtain a property of an object. It is similar to a knowledge graph query, and, in order to be answered,  part of the passage might involve additional sources such as a knowledge graph, or the dataset may have been created using a  knowledge graph, e.g. (\ref{ex:query}) from WikiReading~\cite{hewlett-etal-2016-wikireading}.
    
        \begin{examples}
      \setlength{\itemsep}{0pt}
            \item \textbf{P:} \textit{Cecily Bulstrode (1584-4 August 1609), was a courtier and ... 
            She was the daughter ... }
            \textbf{Q:} \textit{sex or gender}
            \textbf{A:} \textit{female} 
            \label{ex:query}
    \end{examples}

We put datasets with more than one type of question into a separate \textbf{Mixed} category.

\subsection{Passage Type} 

Passages can take the form of  a \textit{one-document} or \textit{multi-document} passage.
They can also be categorised based on the type of reasoning required 
to answer a question: \textit{Simple Evidence} where the answer to a question is clearly presented in the passage, e.g. (\ref{ex:mc}) and (\ref{ex:extractive}), \textit{Multihop Reasoning} with questions requiring that several facts  from different parts of the passage or different documents are combined to obtain the answer, e.g.~(\ref{ex:multihop}) from the HotpotQA~\cite{yang2018hotpotqa}, and \textit{Extended Reasoning} where general knowledge or common sense reasoning is required, e.g. (\ref{ex:cosmos}) from the Cosmos dataset~\cite{huang-etal-2019-cosmos}: 
    \begin{examples}
      \setlength{\itemsep}{0pt}
            \item \textbf{P:} \textit{...2014 S\textbackslash/S is the debut album of South Korean group WINNER. 
            ... 
            WINNER, is a South Korean boy group formed in 2013 by YG Entertainment and debuted in 2014. ...}\\
            \textbf{Q:} \textit{2014 S\textbackslash/S is the debut album of a South Korean boy group that was formed by who?}\\
            \textbf{A:} \textit{YG Entertainment} 
            \label{ex:multihop}
    \end{examples}
    \begin{examples}
          \setlength{\itemsep}{0pt}

        \item \textbf{P:} \textit{I was a little nervous about this today, but they felt fantastic. I think they'll be a very good pair of shoes...} 
        \textbf{Q:} \textit{Why did the writer feel nervous?} 
        \textbf{AC:}
        \textit{(A) None of the above choices.
        (B) Because the shoes felt fantastic.
        (C) Because they were unsure if the shoes would be good quality.
        (D) Because the writer thinks the shoes will be very good.} 
        \textbf{A:} C
        \label{ex:cosmos}
    \end{examples}

\subsection{Conversational MRC} 

    We include \textbf{Conversational} or \textbf{Dialog} datasets in a separate category as they involve a unique combination of passage, question, and answer. 
    The full passage is presented as a conversation and the question should be answered based on previous utterances as illustrated in (\ref{ex:dialog}) from ShARC~\cite{saeidi-etal-2018-interpretation}, where the scenario is an additional part of the passage unique for each dialog. 
    The question and its answer become a part of the passage for the subsequent question.
    \footnote{We include DREAM \cite{sundream2018} in the Multiple-Choice category rather than this category because, even though its passages are in dialog form, the questions are about the dialog but not a part of it.}
    
    \begin{examples}
      \setlength{\itemsep}{0pt}
            \item \textbf{P:} \textit{ Eligibility. You’ll be able to claim the new State Pension if you’re: a man born on or after 6 April 1951, a woman born on or after 6 April 1953} \textbf{Scenario:} \textit{I'm female and I was born in 1966} 
            \textbf{Q:} \textit{Am I able to claim the new State Pension?}
            \textbf{Follow ups:} (1) \textit{Are you a man born on or after 6 April 1951? -- No}
             (2) \textit{Are you a woman born on or after 6 April 1953? -- Yes}
            \textbf{A:} \textit{Yes} 
            \label{ex:dialog}
    \end{examples}



\section{Datasets}\label{sec:datasets}
All datasets and their properties of interest are listed in Table~\ref{table:general}.\footnote{Additional properties and statistics are in  Table~\ref{table:add_prop} in the Appendix \ref{sec:extra}.}
We present the number of questions per dataset (size), the text sources, the method of creation, whether there are a leaderboard and data publicly available, and whether the dataset is \textit{solved}, i.e.~the performance of a MRC system exceeds the reported human performance (also shown).
We will discuss each of these aspects.

\onecolumn

\begin{table*}
\tiny
\begin{longtable}{p{4.7cm}|p{0.8cm}|p{3cm}|p{1.4cm}|p{0.3cm}|p{1.3cm}|p{0.2cm}|p{0.4cm}|p{0.2cm}}

\bf Dataset
& \bf Size (questions) &
\bf Data Source & \bf Q/A Source &\bf LB 
&\bf Human Performance & \bf  Sol-ved 
& \bf TM FW & \bf PAD 
\\ \hline
\endfirsthead
\multicolumn{7}{c}%
{\tablename\ \thetable\ -- \textit{Continued from previous page}} \\
\hline
 \bf Dataset  
 & \bf Size 
 & \bf  Data Source & \bf Q/A Source & \bf  LB 
 & \bf HP & \bf Sol-ved
 & \bf TM FW & \bf PAD  
 \\ \hline
\endhead

\\ \hline \multicolumn{9}{r}{\textit{Continued on next page}} \\
\endfoot

\\ \hline
\endlastfoot
\multicolumn{7}{c}{}\\
\multicolumn{7}{c}{\textbf{Cloze Datasets}}
\\\hline

 \textbf{CNN/Daily Mail} \cite{DBLP:journals/corr/HermannKGEKSB15}& 
    387k/997k&
    CNN/DailyMail & AG &  \textcolor{blue}{\ding{93}} & - & \textcolor{red}{\ding{55}} & -& \textcolor{green}{\ding{51}}\\
    
  \textbf{Children BookTest} \cite{DBLP:journals/corr/HillBCW15}
    & 687k
    & Project Gutenberg& AG & \textcolor{blue}{\ding{93}} & 82  & \textcolor{green}{\ding{51}} & - & \textcolor{green}{\ding{51}}\\

 \textbf{Who Did What}  \cite{onishi-etal-2016-large}
    & 186k 
    & Gigaword & AG & \textcolor{green}{\ding{51}} & 84 & \textcolor{red}{\ding{55}} & - & \textcolor{blue}{\ding{41}}\\
    
 \textbf{BookTest} \cite{DBLP:conf/iclr/BajgarKK17} 
    & 14M
    & Project Gutenberg& AG & \textcolor{red}{\ding{55}} & -  & \textcolor{red}{\ding{55}} & -& \textcolor{red}{\ding{55}}\\    
    
 \textbf{Quasar-S} \cite{DBLP:journals/corr/DhingraMC17}
    & 37k
    & Stack Overflow & AG & \textcolor{red}{\ding{55}}  & 46.8/50.0 & \textcolor{red}{\ding{55}} & -& \textcolor{green}{\ding{51}}\\
    
\textbf{RecipeQA} \cite{yagcioglu-etal-2018-recipeqa} 
    & 9.8k 
    & instructibles.com & AG & \textcolor{green}{\ding{51}} & 73.6& \textcolor{red}{\ding{55}} & -& \textcolor{green}{\ding{51}}\\
    
\textbf{CliCR}  \cite{suster-daelemans-2018-clicr}
    & 105k
    & Clinical Reports & AG & \textcolor{blue}{\ding{93}}  &53.7/45.1 (F1)& \textcolor{red}{\ding{55}} & -& \textcolor{blue}{\ding{41}}\\
    
\textbf{ReCoRD}  \cite{zhang2018record}
    & 121k
    & CNN & AG & \textcolor{green}{\ding{51}}\textcolor{blue}{\ding{93}}  &91.3/91.7 (F1)& \textcolor{green}{\ding{51}} & -& \textcolor{green}{\ding{51}}\\    
    
\textbf{Shmoop}  \cite{chaudhury2019shmoop}
     & 7.2k 
    &  Project Gutenberg &  ER, AG& \textcolor{red}{\ding{55}}  & -& \textcolor{red}{\ding{55}} & -& \textcolor{blue}{\ding{41}}\\
    
\hline

\multicolumn{7}{c}{}\\
\multicolumn{7}{c}{\textbf{Multiple Choice Datasets}}
\\ \hline

 \textbf{MCTest} \cite{richardson2013mctest} 
    &  2k/640
    & Stories & CRW & \textcolor{green}{\ding{51}}\textcolor{blue}{\ding{93}} & 95.3  & \textcolor{red}{\ding{55}}& \textit{what}& \textcolor{green}{\ding{51}}\\
 
 \textbf{WikiQA} \cite{yang-etal-2015-wikiqa}     
    & 3k
    & Wikipedia & UG, CRW& \textcolor{blue}{\ding{93}} & -  & \textcolor{red}{\ding{55}}& \textit{what}& \textcolor{green}{\ding{51}}\\
 
 \textbf{bAbI} \cite{DBLP:journals/corr/WestonBCM15} 
     & 40k
    & AG & AG & \textcolor{blue}{\ding{93}} & 100 & \textcolor{green}{\ding{51}}& \textit{what}& \textcolor{green}{\ding{51}}\\
 
 \textbf{MovieQA} \cite{DBLP:conf/cvpr/TapaswiZSTUF16}
     & 15k
    & Wikipedia &annotators& \textcolor{green}{\ding{51}} & - & \textcolor{red}{\ding{55}} & \textit{what}& \textcolor{blue}{\ding{41}}\\
 
 \textbf{RACE} \cite{lai-etal-2017-race} 
    & 98k
    & ER & experts &\textcolor{green}{\ding{51}}\textcolor{blue}{\ding{93}}& 73.3/94.5  & \textcolor{red}{\ding{55}} & \textit{what}& \textcolor{green}{\ding{51}}\\
    
 \textbf{SciQ} \cite{welbl2017crowdsourcing} 
    & 12k
    & Science Books& CRW &\textcolor{red}{\ding{55}} & 87.8  & \textcolor{red}{\ding{55}} & \textit{what}& \textcolor{green}{\ding{51}}\\
    
 \textbf{MultiRC} \cite{MultiRC2018}
    & 10k
    & reports, News, Wikipedia, ...& CRW &\textcolor{green}{\ding{51}}\textcolor{blue}{\ding{93}} &81.8(F1)& \textcolor{green}{\ding{51}}& \textit{what}& \textcolor{green}{\ding{51}}\\
    
 \textbf{ARC} \cite{ClarketallARC2018}
    & 7.8k
    & ER, Wikipedia, ... & expert  &\textcolor{green}{\ding{51}}\textcolor{blue}{\ding{93}} &-& \textcolor{red}{\ding{55}}& \textit{which}& \textcolor{green}{\ding{51}}\\
    
 \textbf{OpenBookQA} \cite{mihaylov-etal-2018-suit}
    & 6k
    & ER, WorldTree & CRW  &\textcolor{green}{\ding{51}}\textcolor{blue}{\ding{93}} &91.7& \textcolor{red}{\ding{55}}& \textit{-}& \textcolor{green}{\ding{51}}\\
    
 \textbf{MedQA} \cite{DBLP:journals/corr/abs-1802-10279}
    & 235k
    & Medical Books
    & expert&\textcolor{red}{\ding{55}} & -
     & \textcolor{green}{\ding{51}} & -& \textcolor{red}{\ding{55}}\\


 \textbf{MCScript} \cite{ostermann-etal-2018-mcscript} 
    &  14k
    & Scripts, CRW & CRW &\textcolor{red}{\ding{55}}
    & 98.0 & \textcolor{red}{\ding{55}} & \textit{how}& \textcolor{green}{\ding{51}}\\
    
 \textbf{MCScript2.0}  \cite{ostermann-etal-2019-mcscript2} 
    & 20k
    & Scripts, CRW & CRW &\textcolor{red}{\ding{55}}
    & 97.0 & \textcolor{red}{\ding{55}} & \textit{what} & \textcolor{green}{\ding{51}}\\

 \textbf{RACE-C} \cite{pmlr-v101-liang19a} 
    & 14k
    & ER & experts & \textcolor{red}{\ding{55}} & -  & \textcolor{red}{\ding{55}} & \textit{the}& \textcolor{green}{\ding{51}}\\

 \textbf{DREAM}\cite{sundream2018} 
    & 10k
    & ER & experts & \textcolor{green}{\ding{51}} & 98.6 & \textcolor{red}{\ding{55}} & \textit{what}& \textcolor{green}{\ding{51}}\\

 \textbf{Cosmos QA} \cite{huang-etal-2019-cosmos} 
    & 36k
    & Blogs & CRW & \textcolor{green}{\ding{51}} & 94  & \textcolor{red}{\ding{55}} & \textit{what}& \textcolor{green}{\ding{51}}\\
 
 \textbf{ReClor} \cite{Yu2020ReClor:} 
    & 6k
    & ER & experts & \textcolor{green}{\ding{51}} & 63.0  &  \textcolor{green}{\ding{51}}& \textit{which}& \textcolor{green}{\ding{51}}\\

\textbf{QuAIL}  \cite{DBLP:conf/aaai/RogersKDR20}
     & 15k 
    &News, Stories, Fiction, Blogs, UG & CRW, experts& \textcolor{green}{\ding{51}}  & 60.0& \textcolor{red}{\ding{55}} & -& \textcolor{green}{\ding{51}}\\
    
\textbf{QASC}  \cite{khot2020qasc}
     & 10k
    &ER, WorldTree& CRW, ... & \textcolor{green}{\ding{51}}  &93 &  \textcolor{red}{\ding{55}} &  & \textcolor{green}{\ding{51}}\\
    
 \hline
\multicolumn{7}{c}{}\\
\multicolumn{7}{c}{\textbf{Boolean Questions}}\\
 \hline

 \textbf{BoolQ} \cite{clark-etal-2019-boolq} 
    &  16k
    & Wikipedia & UG, CRW&
    \textcolor{green}{\ding{51}}\textcolor{blue}{\ding{93}} & 89  & \textcolor{green}{\ding{51}} & \textit{is}& \textcolor{green}{\ding{51}}\\
    
 \textbf{AmazonYesNo}  \cite{dzendzik-etal-2019-dish}
    & 80k 
    & Reviews  
    & UG     
    & \textcolor{red}{\ding{55}} & - & \textcolor{red}{\ding{55}} & \textit{does}& \textcolor{blue}{\ding{41}}\\
    
 \textbf{PubMedQA}  \cite{pubmedqa}
    & 211k
    & PubMed & CRW & \textcolor{green}{\ding{51}} &78& \textcolor{red}{\ding{55}}& \textit{does} & \textcolor{green}{\ding{51}}\\ \hline

\multicolumn{7}{c}{}\\
\multicolumn{7}{c}{\textbf{Extractive Datasets}}\\
 \hline
 
 
 \textbf{SQuAD} \cite{rajpurkar-etal-2016-squad} 
    & 108k
    & Wikipedia & CRW & \textcolor{green}{\ding{51}}\textcolor{blue}{\ding{93}} &86.8(F1)& \textcolor{green}{\ding{51}}& \textit{what}& \textcolor{green}{\ding{51}}\\
    
 \textbf{SQuAD2.0} \cite{rajpurkar-etal-2018-know} 
    & 151k 
    & Wikipedia & CRW & \textcolor{green}{\ding{51}}\textcolor{blue}{\ding{93}} &89.5(F1)& \textcolor{green}{\ding{51}}& \textit{what}& \textcolor{green}{\ding{51}}\\
 
 \textbf{NewsQA} \cite{trischler-etal-2017-newsqa} 
    & 120k
    & CNN  & CRW &  \textcolor{blue}{\ding{93}} 
    &69.4(F1)& \textcolor{green}{\ding{51}} & \textit{what}& \textcolor{green}{\ding{51}}\\
 
 \textbf{SearchQA} \cite{DBLP:journals/corr/DunnSHGCC17} 
    & 140k
    & CRW, AG & J!Archive&  \textcolor{blue}{\ding{93}} &57.6(F1)& \textcolor{green}{\ding{51}}& \textit{this} & \textcolor{green}{\ding{51}}\\
 \textbf{BiPaR} \cite{jing-etal-2019-bipar} 
    & 14.7k
    & Novels & CRW &  \textcolor{green}{\ding{51}}\textcolor{blue}{\ding{93}} &80.5/91.9(F1)& \textcolor{red}{\ding{55}}& \textit{what} & \textcolor{green}{\ding{51}}\\
    
 \textbf{SubjQA} \cite{bjerva20subjqa} 
    & 10k
    & Reviews & UG, CRW &  \textcolor{blue}{\ding{93}} & - & \textcolor{red}{\ding{55}}& \textit{how} & \textcolor{green}{\ding{51}}\\
    \hline
\multicolumn{7}{c}{}\\
\multicolumn{7}{c}{\textbf{Generative Datasets}}\\
 \hline
 
 \textbf{MS MARCO} \cite{nguyen2016ms}
    & 100k
    & Web documents &  UG, HG & \textcolor{green}{\ding{51}}\textcolor{blue}{\ding{93}}  & -  & \textcolor{red}{\ding{55}}& \textit{what}& \textcolor{green}{\ding{51}}\\
    
 \textbf{LAMBADA} \cite{paperno-etal-2016-lambada}
     & 10k 
    & BookCorpus & CRW, AG& \textcolor{red}{\ding{55}} & -& \textcolor{red}{\ding{55}} & -& \textcolor{green}{\ding{51}}\\

  \textbf{WikiMovies} \cite{miller-etal-2016-key,DBLP:journals/corr/WatanabeDS17}
    & 116k
    & Wikipedia, KG & CRW, AG, KG& \textcolor{red}{\ding{55}}
    &93.9 (hit@1)& \textcolor{red}{\ding{55}} & \textit{what}& \textcolor{green}{\ding{51}}\\
    
 \textbf{WikiSuggest} \cite{choi-etal-2017-coarse} 
    & 3.47M& Wikipedia & CRW, AG& \textcolor{red}{\ding{55}}& - & \textcolor{red}{\ding{55}} & -& \textcolor{red}{\ding{55}}\\
    
 \textbf{TriviaQA} \cite{joshi-etal-2017-triviaqa}
    & 96k
    & Wikipedia, Web docs &Trivia, CRW& \textcolor{green}{\ding{51}}\textcolor{blue}{\ding{93}} & 79.7/75.4 wiki/web& \textcolor{red}{\ding{55}} & \textit{which}& \textcolor{green}{\ding{51}}\\
    
 \textbf{NarrativeQA}  \cite{kocisky-etal-2018-narrativeqa} 
    & 47k
    &Wikipedia, Project Gutenberg, movie, HG& HG & \textcolor{blue}{\ding{93}}&19.7 BLEU4    & \textcolor{green}{\ding{51}} & \textit{what}
    &\textcolor{green}{\ding{51}}\\

\textbf{TweetQA}  \cite{xiong-etal-2019-tweetqa} 
    & 14k
    &News, Twitter, HG& CRW & 
    \textcolor{green}{\ding{51}} & 70.0 BLEU1
    & \textcolor{green}{\ding{51}} & \textit{what}
    &\textcolor{green}{\ding{51}}\\

\hline 
\multicolumn{7}{c}{}\\
\multicolumn{7}{c}{\textbf{Conversational Datasets}}\\
 \hline

 \textbf{ShARC} \cite{saeidi-etal-2018-interpretation}
    & 32k 
    & Legal web sites
    & CRW &\textcolor{green}{\ding{51}} & 93.9  & \textcolor{red}{\ding{55}}& \textit{can}& \textcolor{green}{\ding{51}}\\

 \textbf{CoQA} \cite{reddy-etal-2019-coqa} 
     & 127k
    &Books, News, Wikipedia, ER& CRW& \textcolor{green}{\ding{51}}\textcolor{blue}{\ding{93}} & 88.8  & \textcolor{green}{\ding{51}} & \textit{what}& \textcolor{green}{\ding{51}}\\ 
    
\hline
\multicolumn{7}{c}{}\\
\multicolumn{7}{c}{\textbf{Mixed Datasets}}\\
 \hline 
 
 \textbf{TurkQA} \cite{W13-3213}
    & 54k
    & Wikipedia & CRW & \textcolor{red}{\ding{55}} & -  & \textcolor{red}{\ding{55}}& \textit{what}& \textcolor{green}{\ding{51}}\\

 \textbf{WikiReading}  \cite{hewlett-etal-2016-wikireading}
    & 18.9M
    & Wikipedia & AG, KG & \textcolor{red}{\ding{55}} & - 
    & \textcolor{red}{\ding{55}} & -& \textcolor{green}{\ding{51}}\\

 \textbf{Quasar-T} \cite{DBLP:journals/corr/DhingraMC17}
    & 43k
    & Trivia ClueWeb09& AG &  \textcolor{blue}{\ding{93}} &60.4/60.6 (F1)
    & \textcolor{red}{\ding{55}} & \textit{what}& \textcolor{green}{\ding{51}}\\
    
 \textbf{HotpotQA} \cite{yang2018hotpotqa}
    & 113k
    & Wikipedia & CRW & \textcolor{green}{\ding{51}}\textcolor{blue}{\ding{93}} &96.37(F1)& \textcolor{red}{\ding{55}} & \textit{what}& \textcolor{green}{\ding{51}}\\
    
 \textbf{QAngaroo}  WikiHop \cite{welbl-etal-2018-constructing}
    & 51k
    & Wikipedia & CRW, KG& \textcolor{green}{\ding{51}}\textcolor{blue}{\ding{93}} & 85.0  & \textcolor{red}{\ding{55}} & -& \textcolor{green}{\ding{51}}\\
    
 \textbf{QAngaroo}  MedHop \cite{welbl-etal-2018-constructing}
    & 2.5k
    & Medline abstracts& CRW, KG& \textcolor{green}{\ding{51}} & -  & \textcolor{red}{\ding{55}} & -& \textcolor{green}{\ding{51}}\\
    
 \textbf{QuAC}  \cite{choi-etal-2018-quac}
    & 98k
    & Wikipedia & CRW & \textcolor{green}{\ding{51}}\textcolor{blue}{\ding{93}} &81.1(F1)& \textcolor{red}{\ding{55}}& \textit{what}& \textcolor{green}{\ding{51}}\\
    
 \textbf{DuoRC} \cite{saha-etal-2018-duorc}
    & 86k
    & Wikipedia + IMDB & CRW &     \textcolor{green}{\ding{51}} & -  & \textcolor{red}{\ding{55}} & \textit{who}& \textcolor{green}{\ding{51}}\\

 \textbf{emr QA} \cite{pampari-etal-2018-emrqa}
    & 456k
    & Clinic Records & expert, AG& \textcolor{red}{\ding{55}} & -  & \textcolor{red}{\ding{55}}& \textit{does}& \textcolor{blue}{\ding{41}}\\  

 \textbf{DROP} \cite{dua-etal-2019-drop}
    & 97k
    & Wikipedia & CRW & \textcolor{green}{\ding{51}}
    &96.4(F1)& \textcolor{red}{\ding{55}} & \textit{how} 
    & \textcolor{green}{\ding{51}}\\

 \textbf{NaturalQuestions}  \cite{47761}
    & 323k
    & Wikipedia & UG, CRW& \textcolor{green}{\ding{51}}\textcolor{blue}{\ding{93}} &87/76 L/S(F1)& \textcolor{red}{\ding{55}} & \textit{who}& \textcolor{green}{\ding{51}}\\
    
 \textbf{AmazonQA}  \cite{ijcai2019-694}
    & 570k
    & Review  
    & UG  
    & \textcolor{red}{\ding{55}} & 53.5 & \textcolor{red}{\ding{55}} & \textit{does}
    & \textcolor{green}{\ding{51}}\\
    
\textbf{TyDi} \cite{tydiqa}
    & 11k
    & Wikipedia &  CRW
    & \textcolor{green}{\ding{51}} &54.4(F1)& \textcolor{red}{\ding{55}}& \textit{what}
& \textcolor{green}{\ding{51}}\\ 

 \textbf{R$^3$} \cite{DBLP:journals/corr/abs-2004-01251}
    & 60k
    & Wikipedia & CRW & \textcolor{red}{\ding{55}} & - & \textcolor{red}{\ding{55}} & - & \textcolor{red}{\ding{55}}\\
\textbf{IIRC} \cite{ferguson-etal-2020-iirc}
     & 13k
    & Wikipedia & CRW & \textcolor{blue}{\ding{93}} &  88.4(F1) & \textcolor{red}{\ding{55}} & how & \textcolor{green}{\ding{51}}\\
\hline
\caption{
    Reading comprehension datasets comparison. 
    Where: 
    \textbf{LB} -- leader board available;
    \textbf{Human Performance} --  (expert\textbf{/}non-expert if other not specified): 
        accuracy if other is not specified; 
    \textbf{TMFW} -- the most frequent first word;
    \textbf{PAD} -- publicly available data;
    \textbf{k/M} -- thousands/millions;
    \textbf{CRW} -- crowdsourcing; 
    \textbf{AG} -- automatically generated; 
    \textbf{KG} -- knowledge graph;
    \textbf{ER} -- educational resources;
    \textbf{UG} -- user generated;
    \textbf{HG} -- human generated (UG + annotators, crw, experts);
    \textbf{L/S} -- long/short answer; 
    \textcolor{green}{\ding{51}} -- available/``solved'';
    \textcolor{red}{\ding{55}} -- unavailable/not ``solved'';
    \textcolor{blue}{\ding{93}} -- the leaderboard is hosted at https://paperswithcode.com/;
    \textcolor{blue}{\ding{41}} -- the dataset is available by request.
    This information was last verified in September 2021.
}
\label{table:general}
\end{longtable}
\end{table*}
\twocolumn

\subsection{Data Sources} \label{sec:datasource}
A significant proportion of datasets (23 out of 60) use Wikipedia as a passage source. Eight of those use Wikipedia along with additional sources. 
Other popular sources of text data are:
\begin{enumerate}
\setlength{\itemsep}{-2.5pt}
\item news (NewsQA, MultiRC, ReCoRD, QuAIL, CNN/DailyMail, WhoDidWhat,  CoQA),
\item books, including Project Gutenberg and BookCorpus\footnote{Gutenberg: \url{www.gutenberg.org}
BookCorpus: \url{yknzhu.wixsite.com/mbweb} -- all links last verified (l.v.) 05/2021} \cite{10.1109/ICCV.2015.11} (ChildrenBookTest, BookTest, LAMBADA, BiPaR, partly CoQA, Shmoop, SciQ), 
\item movie scripts (MovieQA, WikiMovies, DuoRC),
and in combination with books (MultiRC and NarrativeQA),
\item medicine: five datasets (CliCR, PubMedQuestions, MedQA, emrQA, QAngaroo MedHop) were created in the medical domain based on clinical reports, medical books, MEDLINE abstracts and PubMed, 
\item exams: RACE, RACE-C, and DREAM use data from English as a Foreign Language examinations,
ReClor from the Graduate Management Admission Test (GMAT) and The Law School Admission Test (LSAT),\footnote{GMAT: \url{www.mba.com/exams/gmat/}, 
LSAT: \url{www.lsac.org/lsat} -- all links l.v. 05/2021} 
MedQA from medical exams, 
and SciQ, ARC, OpenBookQA, and QASC from science exam questions and fact corpora such as WorldTree \cite{jansen-etal-2018-worldtree}.
\end{enumerate}
  Other sources of data include legal resources websites\footnote{For example: \url{www.benefits.gov/}
 \url{www.gov.uk/}, 
 \url{www.usa.gov/} -- all links l.v. 05/2021} (ShARL),  personal narratives  
 from the Spinn3r Blog Dataset~\cite{The_ICWSM_2009_Spinn3r_Dataset} (MCScript, MCScript2.0, CosmosQA),
 StackOverflow (Quasar-S), 
 Quora.com (QuAIL)
 Twitter (TweetQA),\footnote{\newcite{xiong-etal-2019-tweetqa} selected tweets featured in the news.}
reviews from Amazon~\cite{DBLP:conf/icdm/WanM16, DBLP:conf/www/McAuleyY16,DBLP:conf/www/HeM16,ni-etal-2019-justifying},  TripAdvisor,  and Yelp\footnote{\url{https://www.yelp.com/dataset} -- l.v. 09/2021} (AmazonQA, AmazonYesNo, SubjQA), 
 and a cooking website\footnote{\url{www.instructables.com/cooking} -- l.v. 05/2021} (RecipeQA).

\begin{figure*}[ht]
\centering
\includegraphics[width=16cm]{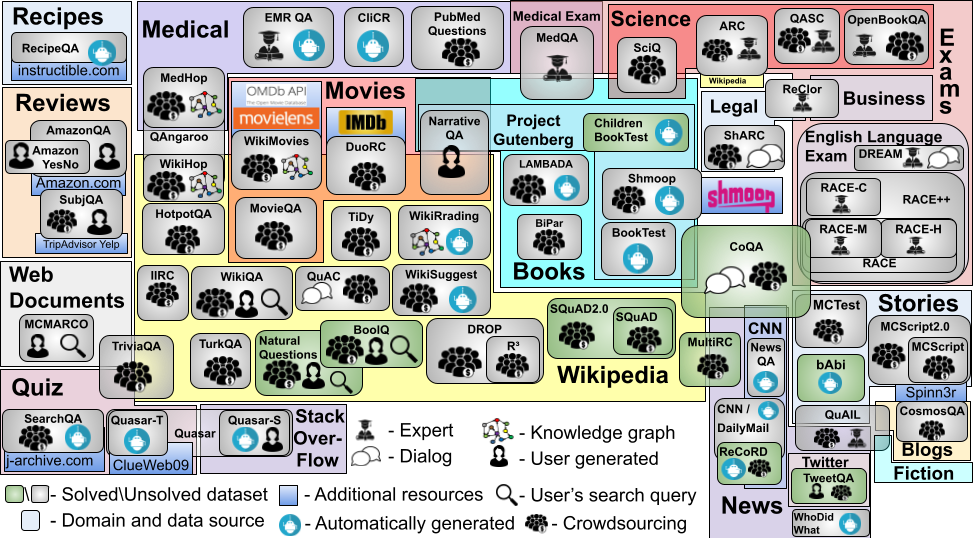}
\caption{Question Answering Reading Comprehension datasets overview. }
\label{fig:overview}
\end{figure*}
Fig.~\ref{fig:overview} shows the domains used by datasets as well as any overlaps between datasets. 
Some datasets share not only text sources but the actual samples. 
SQuAD2.0 extends SQuAD with unanswerable questions. 
AmazonQA and AmazonYesNo overlap in  questions and passages with slightly different processing. 
SubjQA also uses a subset of the same reviews (Movies, Books,
Electronics and Grocery).
BoolQ shares 3k questions and passages with the NaturalQuestions dataset.
The R$^3$ dataset is fully based on DROP with a focus on reasoning. 

\subsection{Dataset Creation}

Rule-based approaches have been used  to automatically obtain questions and passages for the MRC task by generating the sentences (e.g. bAbI) or, in the case of cloze type questions, excluding a word from the context. 
We call those methods \textit{automatically generated} (AG).
Most dataset creation, however, 
involves a human in the loop.
We distinguish three types of people: 
\textit{experts} are professionals in a specific domain;
\textit{crowdworkers} (CRW) are casual workers who normally meet certain criteria (for example a particular level of proficiency in the dataset language) but are not experts in the subject area;
\textit{users} who voluntarily create content based on their personal needs and interests. 

More than half of the datasets (37 out of 60) were created using crowdworkers.
In one scenario, crowdworkers have access to the passage and  must formulate questions based on it. For example, MovieQA, ShaRC, SQuAD, and SQuAD2.0 were created in this way.
In contrast, another scenario involves finding a passage containing the answer for a given question. That works well for datasets where questions are taken from already existing resources such as trivia and quiz questions (TriviaQA, Quasar-T, and SearchQA), 
or using web search queries and results from Google and Bing as a source of questions and passages (BoolQ, NaturalQuestions, MS MARCO). 

In an attempt to avoid word repetition between passages and questions, some datasets used different texts about the same topic as a passage and a source of questions. For example, DuoRC takes descriptions of the same movie from Wikipedia and IMDb.
One description is used as a passage while another is used for creating the questions.
NewsQA uses only a title and a short news article summary as a source for questions while the whole text becomes the passage. Similarly, in NarrativeQA, only the abstracts of the story were used for question creation.
For MCScript and MCScript 2.0, questions and passages were created by different sets of crowdworkers given the same script.

\subsection{Quantitative Analysis}\label{sec:size}

\begin{figure*}[ht]
\centering
\includegraphics[width=15cm]{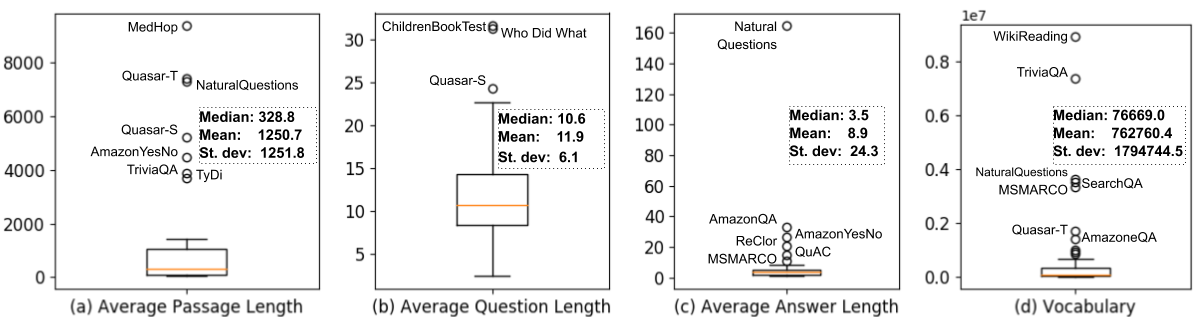}
\caption{
Average length in tokens of (a) passages, (b) questions, (c) answers, and (d) vocabulary size in unique lower-cased lemmas with the \textbf{median}, \textbf{mean} value, and standard deviation (\textbf{St. dev}). Outliers are highlighted.}
\label{fig:stat}
\end{figure*}

Each dataset's size is shown in Table~\ref{table:general}.
About one-third of datasets contain 100k+ questions which makes them suitable for training and/or fine tuning a deep learning model. 
A few datasets contain fewer than 10k samples: MultiRC (9.9k), ARC (7,8k), Shmoop (7.2k), ReClor (6.1k),  OpenBookQA(6k), QAngaroo MedHop (2.5k), WikiQA (2k).
\textcolor{black}{Every dataset has its own structure and data format. We processed all datasets extracting lists of questions, passages, and answers, including answer candidates, and then use the \textit{spaCy}\footnote{\url{spacy.io/api/tokenizer} -- l.v. 03/2020} tokenizer}.

\paragraph{Question/Passage/Answer Length} The graphs in Fig.~\ref{fig:stat} provide more insight into the differences between the datasets in terms of answer, question, and passage length, as well as vocabulary size.
The  outliers are highlighted.\footnote{We use \texttt{matplotlib} for calculation and visualisation: \url{https://matplotlib.org/} -- l.v. 06/2021}
The majority of datasets have a passage length under 1500 tokens with the median being 329 tokens but due to seven outliers, the average number of tokens is 1250 (Fig. \ref{fig:stat} (a)). 
Some datasets (MS MARCO, SearchQA, AmazonYesNo, AmazonQA, MedQA) have a collection of documents as a passage but others contain just a few sentences.\footnote{We consider facts (sentences) from ARC and QASC corpora as different passages. }  
The number of  tokens in a question lies mostly between 5 and 20. 
Two datasets, ChildrenBookTest and WhoDidWhat, have on average more than 30 tokens per question while WikiReading, QAngaroo MedHop, and WikiHope have only 2 -- 3.5 average tokens per question (Fig. \ref{fig:stat} (b)). 
The majority of datasets contain fewer than 8 tokens per answer with the average being 3.5 tokens per answer. The NaturalQuestions is an outlier with average 164 tokens per answer\footnote{We focus on short answers, considering long ones only if the short answer is not available.} (Fig. \ref{fig:stat} (c)). 

\begin{figure*}[ht]
\includegraphics[width=16cm]{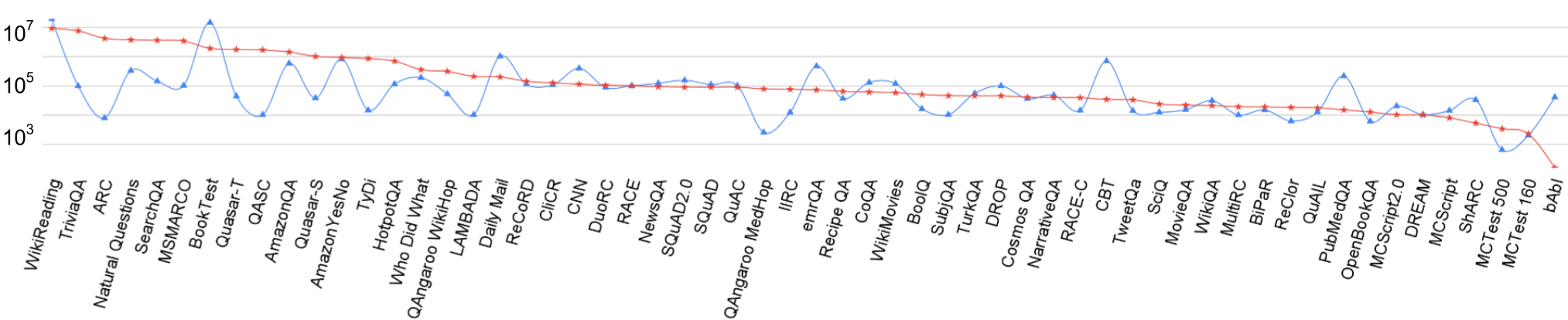}
\caption{The number of questions (blue triangle \textcolor{blue}{\ding{115}}) and unique lower-cased lemmas (vocabulary) (red star \textcolor{red}{\ding{72}}).
}
\label{fig:vocab_correlation}
\end{figure*}

\paragraph{Vocabulary Size} To obtain a vocabulary size we calculate the number of unique lower-cased token lemmas. 
A vocabulary size distribution is presented in Fig. \ref{fig:stat} (d). 
There is a moderate correlation\footnote{
As the data has a non-normal distribution, we calculated the Spearman correlation: coefficient=0.58, p-value=1.3e-05.}
between the number of questions in a dataset and its vocabulary size (see Fig. \ref{fig:vocab_correlation}).\footnote{The values for the BookTest \cite{DBLP:conf/iclr/BajgarKK17} and WhoDidWhat \cite{onishi-etal-2016-large} are taken from the papers.}
WikiReading has the largest number of questions as well as the richest vocabulary. 
bAbI is a synthetic dataset with 40k questions but only 152 lemmas in its vocabulary.

\textbf{Language Detection} We ran a language detector over all datasets using the \texttt{pyenchant} for American and British English, and \texttt{langid} libraries.\footnote{\url{pypi.org/project/pyenchant/} and   \url{github.com/saffsd/langid.py} -- all links l.v. 05/2021.
}
In 37 of the 60 datasets, more than 10\% of the words are reported to be non-English.\footnote{See Appendix~\ref{extra:vocab} and Table~\ref{table:vocab}.} 
We inspected 200 randomly chosen samples from a subset of these.   
For Wikipedia datasets (HotPotQA, QAngoroo WikiHop), 
around 70-75\% of those words are named entities;
10-12\% are specific terms borrowed from other languages such as names of plants, animals, etc.; 
another 8-10\% are foreign words, e.g. the word \textit{``dialetto''}  from HotPotQA \textit{``Bari dialect (dialetto barese) is a dialect of Neapolitan ...''};
about 1.5-3\% are misspelled words and tokenization errors. 
In contrast, for the user-generated dataset, AmazonQA, 67\% are tokenization and spelling errors. 
This aspect of a dataset's vocabulary is useful to bear in mind when, for example, fine-tuning a pre-trained language model which has been trained on less noisy text.
\begin{table}[ht]
\small
 \begin{tabular}{l|c c|c c}
\multicolumn{1}{l}{\multirow{2}{*}{\bf Question}} &
 \multicolumn{2}{|c}{\bf All Questions} & \multicolumn{2}{|c}{\bf Unique Questions}\\
 & \bf Count & \bf \% & \bf Count & \bf \%  \\
 \hline
what&1520443 &22.46\%&1089522 &24.32\%\\
when&139809 &2.07\%&118002 &2.63\%\\
where&157649 &2.33\%&121355 &2.71\%\\
which&280760 &4.15\%&128859 &2.88\%\\
why&98440 &1.45\%&70873 &1.58\%\\
how&473790 &7\%&396188 &8.84\%\\
who/whose&397523 &5.87\%&298260 &6.66\%\\
\hline
boolean&2241553 &33.12\%&1262222 &28.18\%\\
other&1458278 &21.55\%&994551 &22.20\%\\
\end{tabular} 
\caption{First question token frequency across datasets.} \label{table:small_questions} 
\end{table}
\paragraph{First Question Word} A number of datasets come with a breakdown of question types based on the first token~\cite{nguyen2016ms,ostermann-etal-2018-mcscript,ostermann-etal-2019-mcscript2,kocisky-etal-2018-narrativeqa,clark-etal-2019-boolq,xiong-etal-2019-tweetqa, jing-etal-2019-bipar, bjerva20subjqa}.
We inspect the most frequent first word in a dataset's questions excluding cloze-style questions. Table~\ref{table:general} shows the most frequent first word per dataset and  Table~\ref{table:small_questions} shows the same information over all datasets.\footnote{See Appendix~\ref{sec:extra:questions}.} 
The most popular first word is \textit{what} -- 22\% of all questions analysed and over half of the questions in QACS, WikiQA, WikiMovies, MCTest, CosmosQA, and DREAM start with \textit{what}.
 The majority of questions in ReClor (56.5\%) and ARC (28.2\%) start with the  word \textit{which}, and RACE has 23.1\%.  DROP mostly  focused on \textit{how much/many, how old} questions (60.4\%).
 A significant amount of those questions can also be found  IIRC (20\%).
 More than half (56.4\%) of questions in SubjQA are \textit{How} questions.
 DuoRC and BiPar have a significant proportion of \textit{who/whose} questions (39.5\% and 26.4\%). 
\textit{Why}, \textit{When}, and \textit{Where} questions are under-represented -- only 1.5\%, 2\%, and 2.3\% of all questions respectively. Only CosmosQA has a significant proportion (34.2\%) of \textit{Why} questions, MCScript2 (27.9\%) and TyDi (20.5\%) of \textit{When} questions, and bAbI (36.9\%) of \textit{Where} questions.

\section{Evaluation of MRC Datasets}
\subsection{Evaluation Metrics}
\textit{Accuracy} is defined as the ratio of correctly answered questions out of all questions.
For those datasets where the answer should be found or generated (extractive or generative tasks) accuracy is the same as \textit{Exact Match} (EM), implying the  system answer is exactly the same as the gold answer. 
In contrast with selective and boolean tasks, extractive or generative tasks can have ambiguous, incomplete, or redundant answers.
In order to assign credit when the system answer does not exactly match the gold answer, \textit{Precision} and \textit{Recall}, and their harmonic mean, \textit{F1}, can be calculated over words or characters.

Accuracy is used for all boolean, multiple choice, and cloze datasets.\footnote{Except MultiRC as there are multiple correct answers and all of them should be found, and CliCR and ReCoRD which use exact match and F1. This is because even though the task is cloze, the answer should be generated (in case of CliCR) or extracted (ReCoRD).}
For extractive and generative tasks it is common to report EM (accuracy) and F1. For cloze datasets, the metrics depends on the form of answer. If there are options available, accuracy can be calculated. If words have to be generated,  the F1 measure can also be applied.

One can view the MRC task from the perspective of Information Retrieval, providing a ranked list of answers instead of one definitive answer. 
In this case, a Mean Reciprocal Rank~\citep{Craswell2009} (MRR) and Mean Average Precision (MAP) can be used, as well as the accuracy of the top hit (Hits@1) (single answer) over all possible answers (all entities).\footnote{MRR and MAP are used only by \cite{yang-etal-2015-wikiqa} in the WikiQA dataset, as well as precision, recall and F1. \cite{miller-etal-2016-key} in the WikiMovies datasets used the accuracy of the top hit (Hits@1).
} 

All metrics mentioned above work well for well-defined answers but might not reflect performance for generative datasets 
as there could be several alternative ways to answer the same question. 
Some datasets provide more than one gold answer. 
A number of different automatic metrics used in language generation evaluation are also used: 
Bilingual Evaluation Understudy Score (BLEU) \citep{papineni-etal-2002-bleu}, 
Recall Oriented Understudy for Gisting Evaluation (ROUGE-L) \citep{lin-2004-rouge}, and 
Metric for Evaluation of Translation with Explicit ORdering (METEOR) \citep{lavie-agarwal-2007-meteor}.
MSMarco, NarrativeQA, and TweetQA are generative datasets which use these metrics.

\newcite{choi-etal-2018-quac} introduced the human equivalence score (HEQ). 
It measures the percentage of examples where the system F1 matches or exceeds human F1, implying a system’s output is as good as that of an average human. 
There are two variants: HEQ-Q based on questions and HEQ-D based on dialogues.

\subsection{Human Performance}
Human performance figures have been reported for some datasets -- see Table~\ref{table:general}. This is useful in two ways. Firstly, it gives some indication of the difficulty of the questions in the dataset. Contrast, for example, the low human performance score reported for the Quasar and CliCR datasets with the very high scores for DREAM, DROP, and MCScript. Secondly, it provides a comparison point for automatic systems, which may serve to direct researchers to under-studied datasets where the gap between state-of-the-art machine performance and human performance is large, e.g. CliCR (33.9 vs. 53.7), RecipeQA (29.07 vs 73.63), ShaRC (81.2 vs 93.9) and HotpotQA (83.61 vs 96.37).

Although useful, the notion
of human performance is problematic and has to be interpreted with caution. It is usually an average over individuals, whose reading comprehension abilities will vary depending on age, ability to concentrate, interest in, and knowledge of the subject area. Some datasets (CliCR, QUASAR) take the latter into account by distinguishing between expert and non-expert human performance, while RACE distinguishes between crowd-worker and author annotations.
The authors of MedQA, which is based on medical examinations, use a passing mark (of 60\%) as a proxy for human performance. It is important to know this when looking at its ``solved'' status since state-of-the-art accuracy on this dataset is only 75.3\%~\cite{DBLP:journals/corr/abs-1802-10279}.

Finally, \newcite{dunietz-etal-2020-test} call into question the importance of comparing human and machine performance on the MRC task and argue that the questions that MRC systems need to be able to answer are not necessarily the questions that people find difficult to answer. 

\section{Related Work}
Other MRC surveys have been carried out~\cite{DBLP:journals/corr/abs-1907-01118,DBLP:journals/corr/abs-1907-01686,wang2020study,DBLP:journals/corr/abs-1906-03824,DBLP:journals/corr/abs-2001-01582,zeng2020survey,RogersEtAlSurvey2021}. There is room for more than one survey since each survey approaches the field
from a different perspective, varying in the criteria used to decide what datasets to include and what dataset properties to describe. Some focus on MRC \textit{systems} rather than on MRC \textit{datasets}.

%
The surveys closest to ours in coverage are  \newcite{zeng2020survey} 
and \newcite{RogersEtAlSurvey2021}. The survey of \newcite{RogersEtAlSurvey2021} has a wider scope, including datasets for any language, multimodal datasets and datasets which do not fall under the canonical MRC passage/question/answer template.

An emphasis in our survey has been on relatively fine-grained dataset processing. For example, instead of reporting only train/dev/test sizes for each dataset,
we calculate the length of each question, passage and answer. We examine the vocabulary of each dataset, looking at question words and applying language identification tools, in the process discovering question imbalance and noise (HTML tags, misspellings, etc.). We also report data re-use across datasets. 

\section{Concluding Remarks and Recommendations} \label{sec:conclusion}
This paper provides
an overview of English MRC datasets, released up to and including 2020. 
We compare the datasets  by question and answer type, size, data source, creation method, vocabulary, question type, ``solvedness'', and human performance level.
We observe the tendency of moving from smaller datasets towards large collections of questions, and from synthetically generated data through crowdsourcing towards spontaneously created. We also observe a scarcity of \textit{why}, \textit{when}, and \textit{where} questions. 

Gathering and processing the data for this survey was a painstaking task,\footnote{The code is available here:
\url{https://github.com/DariaD/RCZoo} -- l.v. 09/2021}
from which we emerge with some very practical recommendations for future MRC dataset creators. In order to 1) compare to existing datasets, 2) highlight possible limitation for applicable methods, and 3) indicate the computational resources required to process the data, some basic statistics such as average passage/question/answer length, vocabulary size and frequency of question words should be reported; the data itself should be stored in consistent, easy-to-process fashion, ideally with an API provided; any data overlap with existing datasets should be reported; human performance on the dataset should be measured and what it means clearly explained; and finally, if the dataset is for the English language and its design does not differ radically from those surveyed here, 
it is crucial 
to explain why 
this new dataset is needed.

For any future datasets, we suggest a move away from Wikipedia given the volume of existing datasets that are based on it and its use in pre-trained language models such as BERT~\cite{devlin-etal-2019-bert}.
As shown by  \citet{petroni-etal-2019-language}, its use in MRC dataset and pre-training data brings with it the problem that we cannot always determine whether a system's ability to answer a question comes from its comprehension of the relevant passage or from the underlying language model.

The medical domain is well represented in the collection of English MRC datasets, indicating a demand for understanding of this type of text. Datasets may be required for other domains, such as retail, law and government.

Some datasets are designed to test the ability of systems to tell if a question cannot be answered, by including a ``\textit{no answer}'' label. Building upon this, we suggest that datasets be created for the more complex task of
providing qualified answers 
based on different interpretations of the question. 

\section*{Acknowledgements}



This research is partly supported by Science Foundation Ireland in the ADAPT Centre for Digital Content Technology, funded under the SFI Research Centres Programme (Grant 13/RC/2106) and the European Regional Development Fund.
We thank the anonymous reviewers for their helpful feedback.
We also thank Andrew Dunne, Koel Dutta Chowdhury, Valeriia Filimonova, Victoria Serga, Marina Lisuk, Ke Hu, Joachim Wagner, and Alberto Poncelas. 
\bibliography{anthology,emnlp2021}
\bibliographystyle{acl_natbib}

\newpage
\appendix

\newpage

\section{Other Datasets} \label{sec:other}
There are a number of datasets we did not include in our analysis as we focus on the Question Answering Machine Reading Comprehension task. In this section we mention some of these and explain why they are excluded.

CLOTH \cite{xie2018largescale} and Story Cloze Test,  \cite{mostafazadeh-etal-2016-corpus,DBLP:conf/eacl/MostafazadehRLC17} are cloze-style datasets with a word missing from the context and without a specific query. In contrast, cloze question answering  datasets considered in this work have a passage and a separate sentence which can be treated as question with a missing word.

As well as this, we did not include a number of MRC datasets where the story should be completed such as 
        ROCStories~\cite{mostafazadeh-etal-2016-corpus},
        CODAH~\cite{chen-etal-2019-codah},
        SWAG~\cite{zellers-etal-2018-swag}, and
        HellaSWAG~\cite{zellers-etal-2019-hellaswag} because there are no questions.

QBLink~\cite{elgohary-etal-2018-dataset} is technically a MRC QA dataset but for every question there is only the name of a wiki page available. The ``lead in'' information is not enough to answer the question without additional resources. In other words, QBLink is a more general QA dataset, like CommonSenseQA~\cite{talmor-etal-2019-commonsenseqa}.
Another general dataset is MKQA~\cite{DBLP:journals/corr/abs-2007-15207} which contains 260k question-answer pairs in 26 typologically diverse languages.

Textbook Question Answering (TQA)~\cite{8100054} is a multi-modal dataset requiring not only text understanding but also picture processing. 
MCQA is a Multiple Choice Question Answering dataset in English and Chinese based on examination questions introduced as a Shared Task in IJCNLP 2017 by \cite{Shangmin}. The authors do not provide any supportive documents which can be considered as a passage so it is not a reading comprehension task. 


\subsection{Non-English Datasets} \label{sec:nonenglish}
While we focus on English datasets, there are a growing number of MRC datasets in other languages. In this section we will briefly mention some of them. Please see \cite{RogersEtAlSurvey2021} for a more complete list.

\textbf{Chinese datasets:}
DuReader~\cite{he-etal-2018-dureader} is a Chinese RC dataset. It contains a mix of question types based on Baidu Search and Baidu Zhidao.\footnote{\url{zhidao.baidu.com} -- lv. 05/2021} 
BiPaR~\cite{jing-etal-2019-bipar} contains data from a Chinese-English parallel corpora. 
ReCO~\cite{Wang_2020} 
({Re}ading {C}omprehension dataset on {O}pinion) is the largest human-curated Chinese reading comprehension dataset containing 300k questions with \textit{``Yes/No/Unclear''} answers.
Another Chinese dataset is LiveQA~\cite{qianying-etal-2020-liveqa}, which contains  117k multiple-choice questions about sport games with a focus on timeline understanding.

CLUE benchmark~\cite{xu-etal-2020-clue} contains a number of Chinese reading comprehension tasks with different difficulty.  

\textbf{Other Languages:}
The extended version of WikiReading~\cite{byte-level2018kenter} apart from 18M English questions also contains 5M Russian and about 600K Turkish examples.

TyDi~\cite{tydiqa} is a question answering corpus of 11 typologically diverse languages (Arabic, Bengali, Korean, Russian, Telugu, Thai, Finnish, Indonesian, Kiswahili, Japanese, and  English). It contains 200k+ question answer pairs based on the Wikipedia articles in those languages.
MLQA~\cite{lewis-etal-2020-mlqa} contains over 12K question-answer pairs in English and 5K in each of the 6 following languages: Arabic, German, Spanish, Hindi, Vietnamese and Simplified Chinese, with each question-answer instance parallel between 4 other languages on average.

ViMMRC~\cite{nguyen2020pilot} is a multiple-choice questions RC dataset for Vietnamese. It contains 2,783 questions based on a set of 417 texts.

\newcite{hardalov-etal-2019-beyond} created an exam-and-quiz-based dataset for Bulgarian, containing 2,636 multiple-choice questions with additionally extracted context from variety of topics (biology, philosophy, geography, and history).

The RussianSuperGLUE benchmark \cite{shavrina-etal-2020-russiansuperglue} contains several reading comprehension datasets: DaNetQA~\cite{DaNetQA} contains 2.7k boolean questions, MuSeRC and RuCoS~\cite{fenogenova-etal-2020-read} are two datasets with 5k and 87k questions, which require reasoning over multiple sentences and commonsense knowledge to infer the answer.

A number of datasets have been created following the approach of SQuAD:
FQuAD~\cite{dhoffschmidt2020fquad} is a 25,000+ question French Native Reading Comprehension dataset; 
KorQuAD~\cite{lim2019korquad10} has  70,000 original questions in Korean.
The Russian SberQuAD~\citep{SberQuAD} dataset contains about 90K examples.
GermanQuAD~\cite{DBLP:journals/corr/abs-2104-12741} is a German dataset, as the name suggests. It contains almost 14k extractive questions.
All four datasets are based on Wikipedia.  

\textbf{Dataset Translation:} SQuAD has been semi-automatically translated into several other languages such as: 
Korean K-QuAD~\cite{lee-etal-2018-semi};
Italian SQuAD-it~\cite{10.1007/978-3-030-03840-3_29};
Japanese  and French (\cite{DBLP:journals/corr/abs-1809-03275};
Spanish SQuAD-es~\cite{carrino-costajuss-fonollosa:2020:LREC};
Hindi~\cite{10.1145/3359988};
and 
Czech~\cite{mackov2020reading}.


\section{Extra Features and Statistics} \label{sec:extra}
Table~\ref{table:add_prop} contains additional features for each dataset.
\subsection{Calculating the Statistics}
Where possible, we calculated all characteristics based on publicly available data. For those datasets that do not have the test set available, we based our calculations on training and development sets only. 
Those datasets are: 
    AmazonQA, CoQA, DROP, MovieQA, QAngaroo (WikiHop and MedHop), QuAC, ShARC, SQuAD, SQuAD2, TyDi.
Some data we took from original and related papers. 
Those datasets are: 
    BookTest,
    MedQA,
    and R$^3$.

As mentioned in Section~\ref{sec:size}, we processed all datasets in the same way with \textit{spaCy}.
If the data is tokenized or split by sentences we simply join it back using Python \texttt{"~".join(tokens/sentences\_list)}. 
Based on the \textit{spaCy} implementation\footnote{\url{spacy.io} -- l.v. 05/2021} we would not expect significant differences between the originally provided tokenization and the results of the \textit{spaCy} tokenization of the joined tokens.
This ensures consistent processing of all datasets.

There are a few dataset peculiarities that are worth mentioning:

\textbf{ShARC}: there are several scenarios for the same snippet. We consider instance to be a \textit{tree\_id}, and the concatenation of the snippet, scenario and follow up questions with answers as a passage;
    
\textbf{HotpotQA}: we consider a passage to be a concatenation of all supporting facts, and an instance is a title of supporting fact. In this case, there are multiple instances for the same question.

\begin{table*}
\tiny
\begin{tabular}{p{1.8cm}| p{0.4cm} p{0.4cm} p{0.4cm} p{0.5cm} p{0.4cm} p{0.6cm} p{0.5cm} p{0.5cm} | 
p{1cm} p{1cm} p{0.5cm} p{0.8cm} p{0.4cm} p{0.4cm} 
}

 \multicolumn{1}{l}{\multirow{2}{*}{\bf Dataset}} & 
 \multicolumn{8}{c}{\bf Dataset contains} & \multicolumn{6}{c}{\bf Dataset statistics}
 \\ 
 & Yes No & Non-Factoid & Query & Multi Hop & Multi Doc & Dia- logs & No Answer &  Extra Data & \# instances &  \# passages & AVG $Q_{len}$ &  AVG $P_{len}$ &  AVG $A_{len}$
 & Vocab Size \\

\hline



AmazonQA
    &\textcolor{black}{\ding{51}} 
    &\textcolor{black}{\ding{51}} 
    &\textcolor{black}{\ding{55}} 
    &\textcolor{black}{\ding{55}} 
    &\textcolor{black}{\ding{117}} 
    &\textcolor{black}{\ding{55}} 
    &\textcolor{black}{\ding{55}} 
    &\textcolor{black}{\ding{55}} 
    &139,905&830,959&16.6&558.2&32.8&1,395,460\\
    
AmazonYesNo    
    &\textcolor{black}{\ding{51}} 
    &\textcolor{black}{\ding{55}} 
    &\textcolor{black}{\ding{55}} 
    &\textcolor{black}{\ding{117}} 
    &\textcolor{black}{\ding{51}} 
    &\textcolor{black}{\ding{55}} 
    &\textcolor{black}{\ding{117}} 
    &\textcolor{black}{\ding{55}} 
   &40,806&40,806&13.2&4398.2&n/a&864,929\\
   
ARC   
    &\textcolor{black}{\ding{55}} 
    &\textcolor{black}{\ding{51}} 
    &\textcolor{black}{\ding{55}} 
    &\textcolor{black}{\ding{117}} 
    &\textcolor{black}{\ding{117}} 
    &\textcolor{black}{\ding{55}} 
    &\textcolor{black}{\ding{55}} 
    &\textcolor{black}{\ding{55}} 
   &2 & 14.6 &22.5 & 19.4 &4.6 & 4,095,476 \\
 
bAbI
    &\textcolor{black}{\ding{51}} 
    &\textcolor{black}{\ding{55}} 
    &\textcolor{black}{\ding{55}} 
    &\textcolor{black}{\ding{51}} 
    &\textcolor{black}{\ding{55}} 
    &\textcolor{black}{\ding{55}} 
    &\textcolor{black}{\ding{55}} 
    &\textcolor{black}{\ding{55}} 
    &20&1,2534&6.3&67.2&1.1&152\\
    
BiPaR     
    &\textcolor{black}{\ding{117}} 
    &\textcolor{black}{\ding{51}} 
    &\textcolor{black}{\ding{55}} 
    &\textcolor{black}{\ding{51}} 
    &\textcolor{black}{\ding{55}} 
    &\textcolor{black}{\ding{55}} 
    &\textcolor{black}{\ding{55}} 
    &\textcolor{black}{\ding{51}} 
    & 6 & 3.667 & 8.44 & 229.6 & 6.15 & 18597\\
    
BookTest
    &\textcolor{black}{\ding{55}} 
    &\textcolor{black}{\ding{55}} 
    &\textcolor{black}{\ding{55}} 
    &\textcolor{black}{\ding{117}} 
    &\textcolor{black}{\ding{55}} 
    &\textcolor{black}{\ding{117}} 
    &\textcolor{black}{\ding{55}} 
    &\textcolor{black}{\ding{55}} 
    &14,062&14,140,825&-&522&1&1,860,394\\
    
BoolQ
    &\textcolor{black}{\ding{51}} 
    &\textcolor{black}{\ding{55}} 
    &\textcolor{black}{\ding{55}} 
    &\textcolor{black}{\ding{117}} 
    &\textcolor{black}{\ding{55}} 
    &\textcolor{black}{\ding{55}} 
    &\textcolor{black}{\ding{55}} 
    &\textcolor{black}{\ding{55}} 
    &8208&12,697&8.8&109.4&n/a&49,117\\
CBT
    &\textcolor{black}{\ding{55}} 
    &\textcolor{black}{\ding{55}} 
    &\textcolor{black}{\ding{55}} 
    &\textcolor{black}{\ding{55}} 
    &\textcolor{black}{\ding{55}} 
    &\textcolor{black}{\ding{55}} 
    &\textcolor{black}{\ding{55}} 
    &\textcolor{black}{\ding{55}} 
    &108&687,343&30 &440&1&53,628\\
    
CliCR
    &\textcolor{black}{\ding{55}} 
    &\textcolor{black}{\ding{55}} 
    &\textcolor{black}{\ding{55}} 
    &\textcolor{black}{\ding{55}} 
    &\textcolor{black}{\ding{55}} 
    &\textcolor{black}{\ding{55}} 
    &\textcolor{black}{\ding{55}} 
    &\textcolor{black}{\ding{55}} 
    &11,846&11,846&22.6&1411.7&3.4&122,568
    \\
CNN
    &\textcolor{black}{\ding{55}}
    &\textcolor{black}{\ding{55}}
    &\textcolor{black}{\ding{55}}
    &\textcolor{black}{\ding{55}}
    &\textcolor{black}{\ding{55}}
    &\textcolor{black}{\ding{55}}
    &\textcolor{black}{\ding{55}}
    &\textcolor{black}{\ding{55}}
    &n/a&107,122&12.8&708.4&1.4&111,198 \\
    
Daily Mail
    &\textcolor{black}{\ding{55}}
    &\textcolor{black}{\ding{55}}
    &\textcolor{black}{\ding{55}}
    &\textcolor{black}{\ding{55}}
    &\textcolor{black}{\ding{55}}
    &\textcolor{black}{\ding{55}}
    &\textcolor{black}{\ding{55}}
    &\textcolor{black}{\ding{55}}
    &n/a&218,017&14.8&854.4&1.5&197,388 \\

Cosmos QA
    &\textcolor{black}{\ding{55}} 
    &\textcolor{black}{\ding{51}} 
    &\textcolor{black}{\ding{55}} 
    &\textcolor{black}{\ding{51}} 
    &\textcolor{black}{\ding{55}} 
    &\textcolor{black}{\ding{55}} 
    &\textcolor{black}{\ding{55}} 
    &\textcolor{black}{\ding{55}} 
    &35,210&35,210&10.6	&70.4	&8.1	&40,067\\
    
CoQA
    &\textcolor{black}{\ding{51}} 
    &\textcolor{black}{\ding{55}} 
    &\textcolor{black}{\ding{55}} 
    &\textcolor{black}{\ding{117}} 
    &\textcolor{black}{\ding{55}} 
    &\textcolor{black}{\ding{51}} 
    &\textcolor{black}{\ding{51}} 
    &\textcolor{black}{\ding{55}} 
    &n/a&7,699&6.5&328.0&2.9&59,840\\
    
DREAM
    &\textcolor{black}{\ding{117}} 
    &\textcolor{black}{\ding{51}} 
    &\textcolor{black}{\ding{55}} 
    &\textcolor{black}{\ding{51}} 
    &\textcolor{black}{\ding{55}} 
    &\textcolor{black}{\ding{51}} 
    &\textcolor{black}{\ding{55}} 
    &\textcolor{black}{\ding{55}} 
    &6,138&6,444&8.8&86.4&5.3&9,850\\
    
 DROP
    &\textcolor{black}{\ding{117}} 
    &\textcolor{black}{\ding{51}} 
    &\textcolor{black}{\ding{55}} 
    &\textcolor{black}{\ding{51}} 
    &\textcolor{black}{\ding{51}} 
    &\textcolor{black}{\ding{55}} 
    &\textcolor{black}{\ding{55}} 
    &\textcolor{black}{\ding{55}} 
    &n/a&6147&12.2&246.2&4&44,430\\
DuoRC
    &\textcolor{black}{\ding{117}} 
    &\textcolor{black}{\ding{51}} 
    &\textcolor{black}{\ding{55}} 
    &\textcolor{black}{\ding{51}} 
    &\textcolor{black}{\ding{55}} 
    &\textcolor{black}{\ding{55}} 
    &\textcolor{black}{\ding{51}} 
    &\textcolor{black}{\ding{55}} 
    &7,477&7,477&8.6&1,260.9&3.1&119,547\\

emrQA
    &\textcolor{black}{\ding{51}} 
    &\textcolor{black}{\ding{51}} 
    &\textcolor{black}{\ding{117}} 
    &\textcolor{black}{\ding{51}} 
    &\textcolor{black}{\ding{55}} 
    &\textcolor{black}{\ding{55}} 
    &\textcolor{black}{\ding{51}} 
    &\textcolor{black}{\ding{51}} 
    &2427&2,427&7.9&1328.4&2.0&70,837 \\
    
HotpotQA
    &\textcolor{black}{\ding{51}} 
    &\textcolor{black}{\ding{51}} 
    &\textcolor{black}{\ding{55}} 
    &\textcolor{black}{\ding{51}} 
    &\textcolor{black}{\ding{51}} 
    &\textcolor{black}{\ding{55}} 
    &\textcolor{black}{\ding{51}} 
    &\textcolor{black}{\ding{55}} 
    &534,433&105,257&20.0&1100.7&2.4&741,974  \\

IIRC
    &\textcolor{black}{\ding{51}} 
    &\textcolor{black}{\ding{51}} 
    &\textcolor{black}{\ding{55}} 
    &\textcolor{black}{\ding{51}} 
    &\textcolor{black}{\ding{51}} 
    &\textcolor{black}{\ding{55}} 
    &\textcolor{black}{\ding{51}} 
    &\textcolor{black}{\ding{55}} 
    & 5183 & 17,198 & 14.93 & 87.13 & 2.65& 74,777  \\

LAMBADA    
    &\textcolor{black}{\ding{55}} 
    &\textcolor{black}{\ding{55}} 
    &\textcolor{black}{\ding{55}} 
    &\textcolor{black}{\ding{55}} 
    &\textcolor{black}{\ding{55}} 
    &\textcolor{black}{\ding{55}} 
    &\textcolor{black}{\ding{55}} 
    &\textcolor{black}{\ding{55}} 
    &5,325&10,022&15.4&58.5&1&203,918  \\
    
MCScript
    &\textcolor{black}{\ding{51}} 
    &\textcolor{black}{\ding{51}} 
    &\textcolor{black}{\ding{55}} 
    &\textcolor{black}{\ding{55}} 
    &\textcolor{black}{\ding{55}} 
    &\textcolor{black}{\ding{55}} 
    &\textcolor{black}{\ding{55}} 
    &\textcolor{black}{\ding{55}} 
    &110&2,119&6.7&196.0&3.6&7,867\\
    
MCScript2.0
    &\textcolor{black}{\ding{55}} 
    &\textcolor{black}{\ding{51}} 
    &\textcolor{black}{\ding{55}} 
    &\textcolor{black}{\ding{55}} 
    &\textcolor{black}{\ding{55}} 
    &\textcolor{black}{\ding{55}}
    &\textcolor{black}{\ding{117}} 
    &\textcolor{black}{\ding{55}} 
    &200&3,487&8.2&164.4&3.4&11.890\\
    
MCTest 160
    &\textcolor{black}{\ding{51}} 
    &\textcolor{black}{\ding{117}} 
    &\textcolor{black}{\ding{55}} 
    &\textcolor{black}{\ding{117}} 
    &\textcolor{black}{\ding{55}} 
    &\textcolor{black}{\ding{55}} 
    &\textcolor{black}{\ding{55}} 
    &\textcolor{black}{\ding{55}} 
    &160&160&9.2&241.8&3.7&2,246\\
    
MCTest 500
    &\textcolor{black}{\ding{51}} 
    &\textcolor{black}{\ding{117}} 
    &\textcolor{black}{\ding{55}} 
    &\textcolor{black}{\ding{117}} 
    &\textcolor{black}{\ding{55}} 
    &\textcolor{black}{\ding{55}} 
    &\textcolor{black}{\ding{55}} 
    &\textcolor{black}{\ding{55}} 
    &500&500&8.9&251.6&3.8&3,334\\

MedQA
    &\textcolor{black}{\ding{55}} 
    &\textcolor{black}{\ding{51}} 
    &\textcolor{black}{\ding{55}} 
    &\textcolor{black}{\ding{51}} 
    &\textcolor{black}{\ding{51}} 
    &\textcolor{black}{\ding{55}} 
    &\textcolor{black}{\ding{55}} 
    &\textcolor{black}{\ding{55}} 
    &5&243,712&27.4&4.2&43.2&- \\
    
MovieQA
    &\textcolor{black}{\ding{117}} 
    &\textcolor{black}{\ding{51}} 
    &\textcolor{black}{\ding{55}} 
    &\textcolor{black}{\ding{117}} 
    &\textcolor{black}{\ding{55}} 
    &\textcolor{black}{\ding{55}} 
    &\textcolor{black}{\ding{55}} 
    &\textcolor{black}{\ding{51}} 
    &408&408&9.34&727.91&5.6& 21,322\\
    
MSMARCO
    &\textcolor{black}{\ding{51}} 
    &\textcolor{black}{\ding{51}} 
    &\textcolor{black}{\ding{55}} 
    &\textcolor{black}{\ding{51}} 
    &\textcolor{black}{\ding{51}} 
    &\textcolor{black}{\ding{55}} 
    &\textcolor{black}{\ding{51}} 
    &\textcolor{black}{\ding{51}} 
    &n/a&10,087,677&6.5&65.9&11.1&3,324,030\\
    
MultiRC
    &\textcolor{black}{\ding{117}} 
    &\textcolor{black}{\ding{51}} 
    &\textcolor{black}{\ding{55}} 
    &\textcolor{black}{\ding{51}} 
    &\textcolor{black}{\ding{51}} 
    &\textcolor{black}{\ding{55}} 
    &\textcolor{black}{\ding{55}} 
    &\textcolor{black}{\ding{55}} 
    &871&871&4.8&92.4&5.5& 23,331\\
    
NarrativeQA    
    &\textcolor{black}{\ding{55}} 
    &\textcolor{black}{\ding{51}} 
    &\textcolor{black}{\ding{55}} 
    &\textcolor{black}{\ding{51}} 
    &\textcolor{black}{\ding{117}} 
    &\textcolor{black}{\ding{51}} 
    &\textcolor{black}{\ding{55}} 
    &\textcolor{black}{\ding{55}} 
    &1,572&1,572&9.9&673.9&4.8&38,870\\
    
Natural Questions
    &\textcolor{black}{\ding{51}} 
    &\textcolor{black}{\ding{51}} 
    &\textcolor{black}{\ding{55}} 
    &\textcolor{black}{\ding{55}} 
    &\textcolor{black}{\ding{55}} 
    &\textcolor{black}{\ding{55}} 
    &\textcolor{black}{\ding{51}} 
    &\textcolor{black}{\ding{55}} 
    &109,715&315,203&9.36&7312.13&164.56&3,635,821\\
    
NewsQA   
    &\textcolor{black}{\ding{117}} 
    &\textcolor{black}{\ding{51}} 
    &\textcolor{black}{\ding{55}} 
    &\textcolor{black}{\ding{51}} 
    &\textcolor{black}{\ding{55}} 
    &\textcolor{black}{\ding{55}} 
    &\textcolor{black}{\ding{51}} 
    &\textcolor{black}{\ding{55}} 
    &12,744&12,744&7.8&749.2&5.0&90,854 \\

OpenBookQA   
    &\textcolor{black}{\ding{55}} 
    &\textcolor{black}{\ding{51}} 
    &\textcolor{black}{\ding{55}} 
    &\textcolor{black}{\ding{117}} 
    &\textcolor{black}{\ding{117}} 
    &\textcolor{black}{\ding{55}} 
    &\textcolor{black}{\ding{55}} 
    &\textcolor{black}{\ding{55}} 
   &n/a & 5957 &11.44 & 9.39 &2.9 & 12430\\

PubMedQA
    &\textcolor{black}{\ding{51}} 
    &\textcolor{black}{\ding{55}} 
    &\textcolor{black}{\ding{55}} 
    &\textcolor{black}{\ding{51}} 
    &\textcolor{black}{\ding{55}} 
    &\textcolor{black}{\ding{55}} 
    &\textcolor{black}{\ding{117}} 
    &\textcolor{black}{\ding{55}} 
    &n/a&3,358& 3 &15.1&73.8&14,751\\
    
QAngaroo WikiHop  
    &\textcolor{black}{\ding{55}} 
    &\textcolor{black}{\ding{55}} 
    &\textcolor{black}{\ding{51}} 
    &\textcolor{black}{\ding{51}} 
    &\textcolor{black}{\ding{51}} 
    &\textcolor{black}{\ding{55}} 
    &\textcolor{black}{\ding{55}} 
    &\textcolor{black}{\ding{55}} 
    &n/a&48,867&3.5&1381&1.8&304,322\\
    
QAngaroo MedHop
    &\textcolor{black}{\ding{55}} 
    &\textcolor{black}{\ding{55}} 
    &\textcolor{black}{\ding{51}} 
    &\textcolor{black}{\ding{51}} 
    &\textcolor{black}{\ding{51}} 
    &\textcolor{black}{\ding{55}} 
    &\textcolor{black}{\ding{55}} 
    &\textcolor{black}{\ding{55}} 
    &n/a&1962&3&9366.7&1&76,954\\
    
QASC   
    &\textcolor{black}{\ding{55}} 
    &\textcolor{black}{\ding{51}} 
    &\textcolor{black}{\ding{55}} 
    &\textcolor{black}{\ding{51}} 
    &\textcolor{black}{\ding{117}} 
    &\textcolor{black}{\ding{55}} 
    &\textcolor{black}{\ding{55}} 
    &\textcolor{black}{\ding{55}} 
   & n/a & 17M & 9.6 & 12.4 & 1.67 & 1,637,960\\

QuAC    
    &\textcolor{black}{\ding{51}} 
    &\textcolor{black}{\ding{51}} 
    &\textcolor{black}{\ding{55}} 
    &\textcolor{black}{\ding{51}} 
    &\textcolor{black}{\ding{55}} 
    &\textcolor{black}{\ding{51}} 
    &\textcolor{black}{\ding{51}} 
    &\textcolor{black}{\ding{55}} 
    &8853&13,594 
    &5.6&401 
    &14.1& 99,912
    \\

 QuAIL
    &\textcolor{black}{\ding{55}} 
    &\textcolor{black}{\ding{51}} 
    &\textcolor{black}{\ding{55}} 
    &\textcolor{black}{\ding{51}} 
    &\textcolor{black}{\ding{55}} 
    &\textcolor{black}{\ding{55}} 
    &\textcolor{black}{\ding{51}} 
    &\textcolor{black}{\ding{55}} 
    &680&680&9.70&388.29 
    &4.36& 17271\\

Quasar-S
    &\textcolor{black}{\ding{117}} 
    &\textcolor{black}{\ding{55}} 
    &\textcolor{black}{\ding{55}} 
    &\textcolor{black}{\ding{55}} 
    &\textcolor{black}{\ding{55}} 
    &\textcolor{black}{\ding{55}} 
    &\textcolor{black}{\ding{55}} 
    &\textcolor{black}{\ding{55}} 
    &n/a&37,362&24.3&(S)1995.9 (L)5210.1&1.5&(S)660,425 (L)987,380\\
    
Quasar-T
    &\textcolor{black}{\ding{55}} 
    &\textcolor{black}{\ding{55}} 
    &\textcolor{black}{\ding{55}} 
    &\textcolor{black}{\ding{55}} 
    &\textcolor{black}{\ding{55}} 
    &\textcolor{black}{\ding{55}} 
    &\textcolor{black}{\ding{55}} 
    &\textcolor{black}{\ding{55}} 
    &n/a&43,012&11.1&(S)2256.2 (L)7372.6&1.9&(S)1,021,823 (L)2,019,336  \\

RACE
    &\textcolor{black}{\ding{55}} 
    &\textcolor{black}{\ding{51}} 
    &\textcolor{black}{\ding{55}} 
    &\textcolor{black}{\ding{51}} 
    &\textcolor{black}{\ding{55}} 
    &\textcolor{black}{\ding{55}} 
    &\textcolor{black}{\ding{55}} 
    &\textcolor{black}{\ding{55}} 
    &n/a&27,933&12.0&329.5&6.3&98,482\\
    
RACE-C
    &\textcolor{black}{\ding{55}} 
    &\textcolor{black}{\ding{55}} 
    &\textcolor{black}{\ding{55}} 
    &\textcolor{black}{\ding{51}} 
    &\textcolor{black}{\ding{55}} 
    &\textcolor{black}{\ding{55}} 
    &\textcolor{black}{\ding{55}} 
    &\textcolor{black}{\ding{55}} 
    &n/a&2,708&13.8&423.8&7.4&38,399\\
    
Recipe QA
    &\textcolor{black}{\ding{55}} 
    &\textcolor{black}{\ding{55}} 
    &\textcolor{black}{\ding{55}} 
    &\textcolor{black}{\ding{55}} 
    &\textcolor{black}{\ding{55}} 
    &\textcolor{black}{\ding{55}} 
    &\textcolor{black}{\ding{55}} 
    &\textcolor{black}{\ding{51}} 
    &n/a&9,761&10.8&580.0&3.3&62,938\\
    
ReClor
    &\textcolor{black}{\ding{55}} 
    &\textcolor{black}{\ding{51}} 
    &\textcolor{black}{\ding{55}} 
    &\textcolor{black}{\ding{51}} 
    &\textcolor{black}{\ding{55}} 
    &\textcolor{black}{\ding{55}} 
    &\textcolor{black}{\ding{55}} 
    &\textcolor{black}{\ding{55}} 
    &n/a&6,138&17.0&73.6&20.6&17,865 \\
    
ReCoRD
    &\textcolor{black}{\ding{55}} 
    &\textcolor{black}{\ding{55}} 
    &\textcolor{black}{\ding{55}} 
    &\textcolor{black}{\ding{51}} 
    &\textcolor{black}{\ding{55}} 
    &\textcolor{black}{\ding{55}} 
    &\textcolor{black}{\ding{55}} 
    &\textcolor{black}{\ding{55}} 
    &n/a&73190&24.72&193.64&1.5&139724 \\
    
SciQ
    &\textcolor{black}{\ding{55}} 
    &\textcolor{black}{\ding{55}} 
    &\textcolor{black}{\ding{55}} 
    &\textcolor{black}{\ding{55}} 
    &\textcolor{black}{\ding{55}} 
    &\textcolor{black}{\ding{55}} 
    &\textcolor{black}{\ding{55}} 
    &\textcolor{black}{\ding{51}} 
    &n/a&12,252&14.6&87.1&1.5&23,320\\
    
SearchQA    
    &\textcolor{black}{\ding{55}} 
    &\textcolor{black}{\ding{51}} 
    &\textcolor{black}{\ding{55}} 
    &\textcolor{black}{\ding{51}} 
    &\textcolor{black}{\ding{51}} 
    &\textcolor{black}{\ding{55}} 
    &\textcolor{black}{\ding{55}} 
    &\textcolor{black}{\ding{55}} 
    &27,995&13,796,295&16.7&58.7&2.1&3,506,501\\ 
    
ShARC
    &\textcolor{black}{\ding{51}} 
    &\textcolor{black}{\ding{55}} 
    &\textcolor{black}{\ding{55}} 
    &\textcolor{black}{\ding{117}} 
    &\textcolor{black}{\ding{55}} 
    &\textcolor{black}{\ding{51}} 
    &\textcolor{black}{\ding{55}} 
    &\textcolor{black}{\ding{55}} 
    &697&24,160&8.6&87.2&4.0&5,231 
    \\

SQuAD 
    &\textcolor{black}{\ding{55}} 
    &\textcolor{black}{\ding{51}} 
    &\textcolor{black}{\ding{55}} 
    &\textcolor{black}{\ding{55}} 
    &\textcolor{black}{\ding{55}} 
    &\textcolor{black}{\ding{55}} 
    &\textcolor{black}{\ding{55}} 
    &\textcolor{black}{\ding{55}} 
    &490&20,963&11.4&137.1&3.5&87,765\\
    
SQuAD2.0    
    &\textcolor{black}{\ding{55}} 
    &\textcolor{black}{\ding{51}} 
    &\textcolor{black}{\ding{55}} 
    &\textcolor{black}{\ding{55}} 
    &\textcolor{black}{\ding{55}} 
    &\textcolor{black}{\ding{55}} 
    &\textcolor{black}{\ding{51}} 
    &\textcolor{black}{\ding{55}} 
    &477&20,239&11.2&137.0&3.5&88,081\\

SubjQA    
    &\textcolor{black}{\ding{51}} 
    &\textcolor{black}{\ding{51}} 
    &\textcolor{black}{\ding{55}} 
    &\textcolor{black}{\ding{55}} 
    &\textcolor{black}{\ding{55}} 
    &\textcolor{black}{\ding{55}} 
    &\textcolor{black}{\ding{51}} 
    &\textcolor{black}{\ding{55}} 
    & 6 & 10093 & 6.56 & 274.27 & 3.68 & 45,636\\

TriviaQA    
    &\textcolor{black}{\ding{55}} 
    &\textcolor{black}{\ding{117}} 
    &\textcolor{black}{\ding{55}} 
    &\textcolor{black}{\ding{51}} 
    &\textcolor{black}{\ding{51}} 
    &\textcolor{black}{\ding{55}} 
    &\textcolor{black}{\ding{51}} 
    &\textcolor{black}{\ding{55}} 
    &n/a&801,194&16.4&3867.6&2.3&7,366,586\\
    
TurkQA
    &\textcolor{black}{\ding{51}} 
    &\textcolor{black}{\ding{51}} 
    &\textcolor{black}{\ding{55}} 
    &\textcolor{black}{\ding{55}} 
    &\textcolor{black}{\ding{55}} 
    &\textcolor{black}{\ding{55}} 
    &\textcolor{black}{\ding{55}} 
    &\textcolor{black}{\ding{55}} 
    &n/a&13,425&10.3&41.6&2.9&44,677\\

TweetQA
    &\textcolor{black}{\ding{55}} 
    &\textcolor{black}{\ding{51}} 
    &\textcolor{black}{\ding{55}} 
    &\textcolor{black}{\ding{55}} 
    &\textcolor{black}{\ding{55}} 
    &\textcolor{black}{\ding{55}} 
    &\textcolor{black}{\ding{55}} 
    &\textcolor{black}{\ding{55}} 
    &n/a&13757&8.02&31.93&2.70&32542\\

TyDi    
    &\textcolor{black}{\ding{51}} 
    &\textcolor{black}{\ding{55}} 
    &\textcolor{black}{\ding{55}} 
    &\textcolor{black}{\ding{55}} 
    &\textcolor{black}{\ding{55}} 
    &\textcolor{black}{\ding{55}} 
    &\textcolor{black}{\ding{51}} 
    &\textcolor{black}{\ding{51}} 
    &n/a&14,378&8.3&3,694.2&4.6&848,524\\

Who Did What
    &\textcolor{black}{\ding{55}} 
    &\textcolor{black}{\ding{55}} 
    &\textcolor{black}{\ding{55}} 
    &\textcolor{black}{\ding{55}} 
    &\textcolor{black}{\ding{55}} 
    &\textcolor{black}{\ding{55}} 
    &\textcolor{black}{\ding{55}} 
    &\textcolor{black}{\ding{55}} 
    &n/a&205,978&31.2&N/A&2.1&347,406\\
    
WikiMovies     
    &\textcolor{black}{\ding{55}} 
    &\textcolor{black}{\ding{51}} 
    &\textcolor{black}{\ding{51}} 
    &\textcolor{black}{\ding{117}} 
    &\textcolor{black}{\ding{51}} 
    &\textcolor{black}{\ding{55}} 
    &\textcolor{black}{\ding{55}} 
    &\textcolor{black}{\ding{51}} 
    &n/a&186,444&8.7&77.9&6.8&56,893\\
    
WikiQA
    &\textcolor{black}{\ding{55}} 
    &\textcolor{black}{\ding{55}} 
    &\textcolor{black}{\ding{55}} 
    &\textcolor{black}{\ding{55}} 
    &\textcolor{black}{\ding{55}} 
    &\textcolor{black}{\ding{55}} 
    &\textcolor{black}{\ding{51}} 
    &\textcolor{black}{\ding{55}} 
    &n/a&1,242&6.5&252.6&n/a&20,686\\
    
WikiReading    
    &\textcolor{black}{\ding{55}} 
    &\textcolor{black}{\ding{55}} 
    &\textcolor{black}{\ding{51}} 
    &\textcolor{black}{\ding{51}} 
    &\textcolor{black}{\ding{55}} 
    &\textcolor{black}{\ding{55}} 
    &\textcolor{black}{\ding{51}} 
    &\textcolor{black}{\ding{55}} 
    &4,313,786&18,807,888&2.35	&569.0&2.2&8,928,645\\

\hline
\end{tabular}
\caption[Additional properties of Datasets]{Datasets in alphabetical order and additional properties.  Where extra data means the English RC task is only one part of bigger dataset with additional resources such as images or video, or there is an availability of resources in other languages.    
\ding{51}  -- presented;
\ding{117} -- presented in a limited form;
\ding{55}  -- not presented.
}
\label{table:add_prop}

\end{table*}

\textbf{TyDi}: we calculated the statistic for joined data from the English subset for both the Minimal answer span task and the Gold passage task for training and development sets.
    
\textbf{WikiQA}: we consider a passage to be the concatenation of all sentences, we did calculations based on publicly available data and code from the github  page;\footnote{\url{github.com/RaRe-Technologies/gensim-data/issues/31} -- l.v. 05/2021}
    
\textbf{TriviaQA}: to obtain the data we modified the script\footnote{\url{github.com/mandarjoshi90/triviaqa} -- l.v. 05/2021} provided by the authors;
    
\textbf{MSMARCO}: we consider every passage separately, so in this case, there are multiple passages for one question.

\textbf{Who Did What}: we looked into relaxed setting. We do not have a licence to get Gigaword data so we calculated only the average length of the questions and answers. The vocabulary size is provided by the original paper \cite{onishi-etal-2016-large}.

\subsection{Vocabulary}\label{extra:vocab}
Table \ref{table:vocab} contains vocabulary analysis per dataset.

\begin{center}
\begin{table*}
\small
\begin{tabular}{p{2.6cm}|p{2.2cm}p{2cm}p{2.8cm}p{1.9cm} p{1.9cm}}

 \bf Dataset&\bf English Words&\bf Numbers&\bf Not-English Words&\bf Not ASCII&\bf Web Links \\\hline

\hline
AmazonQA	&\small{1065795} (76.4\%)	&38323 (2.7\%)	&235019 (16.8\%)	&6240 (0.4\%)	&\small{49765} (3.6\%)\\  
\small{AmazonYesNo}	&736037 (81.3\%)	&17931 (2.0\%)	&144761 (16.0\%)	&45 (0.0\%)	&6345 (0.7\%)\\  
bAbI	&145 (95.4\%)	&0(0\%)	&7 (4.6\%)	&0(0\%)	&0(0\%)\\  
BiPaR	&17118 (92.0\%)	&111 (0.6\%)	&712 (3.8\%)	&654 (3.5\%)	&2 (0.0\%)\\  
BoolQ	&36940 (75.2\%)	&3050 (6.2\%)	&7081 (14.4\%)	&2007 (4.1\%)	&41 (0.1\%)\\  
CBTest	&29630 (88.4\%)	&167 (0.5\%)	&3651 (10.9\%)	&58 (0.2\%)	&0(0\%)\\  
CNN	&75523 (67.9\%)	&6290 (5.7\%)	&27250 (24.5\%)	&726 (0.7\%)	&1408 (1.3\%)\\  
CliCR	&82981 (67.7\%)	&7798 (6.4\%)	&30809 (25.1\%)	&890 (0.7\%)	&85 (0.1\%)\\  
CoQA	&45112 (75.4\%)	&2605 (4.4\%)	&10270 (17.2\%)	&1748 (2.9\%)	&93 (0.2\%)\\  
CosmosQA	&34466 (86.0\%)	&934 (2.3\%)	&4617 (11.5\%)	&6 (0.0\%)	&42 (0.1\%)\\  
DREAM	&8653 (87.8\%)	&711 (7.2\%)	&469 (4.8\%)	&11 (0.1\%)	&2 (0.0\%)\\  
DROP	&27458 (61.8\%)	&7564 (17.0\%)	&7545 (17.0\%)	&1840 (4.1\%)	&13 (0.0\%)\\  
DailyMail	&130062 (65.9\%)	&13919 (7.1\%)	&49752 (25.2\%)	&1457 (0.7\%)	&2197 (1.1\%)\\  
DuoRC	&73800 (72.5\%)	&1235 (1.2\%)	&22937 (22.5\%)	&3715 (3.6\%)	&33 (0.0\%)\\  
emrQA	&48174 (68.0\%)	&\small{12287} (17.3\%)	&10060 (14.2\%)	&2 (0.0\%)	&0(0\%)\\  
HotPotQA	&341142 (50.2\%)	&29140 (4.3\%)	&199911 (29.4\%)	&\small{107605}(15.8\%)	&1901 (0.3\%)\\  
IIRC	&55755 (74.6\%)	&3564 (4.8\%)	&11532 (15.4\%)	&3912 (5.2\%)	&14 (0.0\%)\\  
LAMBADA	&144310 (70.8\%)	&4828 (2.4\%)	&49745 (24.4\%)	&2846 (1.4\%)	&2186 (1.1\%)\\  
MCScript	&7544 (95.9\%)	&101 (1.3\%)	&198 (2.5\%)	&15 (0.2\%)	&6 (0.1\%)\\  
MCScript2	&9467 (94.4\%)	&138 (1.4\%)	&395 (3.9\%)	&17 (0.2\%)	&12 (0.1\%)\\  
MCTest 160	&2135 (95.1\%)	&31 (1.4\%)	&74 (3.3\%)	&1 (0.0\%)	&0(0\%)\\  
MCTest 500	&3145 (94.3\%)	&35 (1.0\%)	&147 (4.4\%)	&1 (0.0\%)	&0(0\%)\\  
MSMARCO	&\small{2046615} (61.6\%)	&\small{261290} (7.9\%)	&703298 (21.2\%)	&\small{246936} (7.4\%)	&\small{65825} (2.0\%)\\  
MovieQA	&18166 (85.2\%)	&385 (1.8\%)	&2768 (13.0\%)	&1 (0.0\%)	&0(0\%)\\  
MultiRC	&16034 (84.9\%)	&896 (4.7\%)	&1821 (9.6\%)	&106 (0.6\%)	&14 (0.1\%)\\  
NarrativeQA	&31058 (79.9\%)	&631 (1.6\%)	&6213 (16.0\%)	&927 (2.4\%)	&1 (0.0\%)\\  
\small{NaturalQuestions}	&\small{1177894} (32.4\%)	&\small{891487} (24.5\%)	&757428 (20.8\%)	&\small{364341}(10.0\%)	&\small{444670}(12.2\%)\\  
NewsQA	&65487 (72.1\%)	&4316 (4.7\%)	&19370 (21.3\%)	&716 (0.8\%)	&950 (1.0\%)\\  
OpenBookQA	&8370 (97.3\%)	&94 (1.1\%)	&136 (1.6\%)	&0(0\%)	&0(0\%)\\  
PubMedQA	&11139 (75.4\%)	&2531 (17.1\%)	&941 (6.4\%)	&148 (1.0\%)	&1 (0.0\%)\\  
\small{QAngoroo MedHop}	&59186 (77.2\%)	&4858 (6.3\%)	&10877 (14.2\%)	&1722 (2.2\%)	&26 (0.0\%)\\  
\small{QAngoroo WikiHop}	&173858 (57.1\%)	&22415 (7.4\%)	&93948 (30.9\%)	&\small{13860} (4.6\%)	&345 (0.1\%)\\  
QuAC	&63683 (72.6\%)	&3499 (4.0\%)	&20315 (23.2\%)	&101 (0.1\%)	&107 (0.1\%)\\  
Quasar-S	&622534 (63.0\%)	&\small{109403} (11.1\%)	&210401 (21.3\%)	&1 (0.0\%)	&\small{45475} (4.6\%)\\  
Quasar-T	&941480 (55.5\%)	&167738(9.9\%)	&479864 (28.3\%)	&1 (0.0\%)	&\small{107374}(6.3\%)\\  
RACE-C	&30697 (79.9\%)	&1248 (3.3\%)	&3988 (10.4\%)	&2420 (6.3\%)	&30 (0.1\%)\\  
RACE	&75342 (76.5\%)	&6277 (6.4\%)	&15863 (16.1\%)	&1 (0.0\%)	&889 (0.9\%)\\  
ReClor	&16364 (91.6\%)	&326 (1.8\%)	&1174 (6.6\%)	&1 (0.0\%)	&0(0\%)\\  
RecipeQA	&48929 (77.0\%)	&1031 (1.6\%)	&10560 (16.6\%)	&1181 (1.9\%)	&835 (1.3\%)\\  
SearchQA	&\small{2129356} (60.7\%)	&\small{313517} (8.9\%)	&957977 (27.3\%)	&392 (0.0\%)	&\small{105207} (3.0\%)\\  
ShaRC	&4703 (90.6\%)	&303 (5.8\%)	&161 (3.1\%)	&15 (0.3\%)	&1 (0.0\%)\\  
SQuAD	&58444 (66.6\%)	&5708 (6.5\%)	&16827 (19.2\%)	&6706 (7.6\%)	&55 (0.1\%)\\  
SQuAD2	&58793 (66.8\%)	&5724 (6.5\%)	&16935 (19.2\%)	&6548 (7.4\%)	&54 (0.1\%)\\  
SubjQA	&40624 (89.0\%)	&1118 (2.4\%)	&3808 (8.3\%)	&33 (0.1\%)	&53 (0.1\%)\\  
TriviaQA	&\small{3269469} (44.3\%)	&\small{421543} (5.7\%)	&\small{1566735} (21.2\%)	&\small{1806003}(24.5\%)	&\small{293086} (4.0\%)\\  
TurkQA	&32225 (72.1\%)	&1660 (3.7\%)	&10778 (24.1\%)	&1 (0.0\%)	&25 (0.1\%)\\  
TyDi	&532336 (61.8\%)	&31785 (3.7\%)	&184915 (21.5\%)	&\small{83113}(9.6\%)	&828 (0.1\%)\\  
WhoDidWhat	&79056 (63.5\%)	&2658 (2.1\%)	&42670 (34.3\%)	&40 (0.0\%)	&52 (0.0\%)\\  
WikiMovies	&39249 (69.0\%)	&447 (0.8\%)	&15310 (26.9\%)	&1880 (3.3\%)	&3 (0.0\%)\\  
WikiQA	&17074 (82.5\%)	&1081 (5.2\%)	&2041 (9.9\%)	&477 (2.3\%)	&12 (0.1\%)\\  
WikiReading	&\small{3431134}(38.4\%)	&\small{823603}(9.2\%)	&\small{2777734}(31.1\%)	&\small{1801580}(20.2\%)	&\small{94594}(1.1\%)\\  

\end{tabular}

\caption{Types of lemmas in dataset's vocabulary in percentage listed in decreasing order according to the vocabulary size.}
\label{table:vocab} 

\end{table*}
\end{center}

\subsection{Questions}\label{sec:extra:questions}
Some questions could be formulated with a question word inside, for example: \textit{"About how much does each box of folders weigh?"} or \textit{"According to the narrator, what may be true about their employer?"}. 
We analyse 6.8M questions excluding all cloze datasets (ChildrenBookTest, CNN/DailyMail, WhoDidWhat, CliCR, LAMBADA, RecipeQA, Quasar-S, some cloze style questions from MS MARCO, DREAM, Quasar-T, RACE, RACE-C, SearchQA, TriviaQA, emrQA) (there are all together approximately 2.5M cloze questions) and WikiReading, WikiHop, and MedHop (almost 19 million questions-queries),  as the queries are not formulated in question form. As mentioned in section \ref{sec:datasource} some datasets shared the questions and some datasets have the same questions asked more than once within a different context (for example, question \textit{"Where is Daniel?"} asked 2007 times in bAbI), or same questions asked with different answer options (for example in CosmosQA dataset). 


To separate boolean questions we used the same list of words as \newcite{clark-etal-2019-boolq}: \textit{“did”, “do”, “does”, “is”, “are”, “was”,
“were”, “have”, “has”, “can”, “could”, “will”, “would”}. 
Apart from datasets which contain only yes/no/maybe questions a significant portion of boolean questions are in
ShaRC (85.4\%),
emrQA (74.0 \%)
AmazonQA (55.3\%),
QuAC (36.6\%),
MCSCript (28.6\%),
TurcQA (25.7\%),
bAbI (25.0 \%)
and CoQA (20.7\%).

Almost a third of all questions and more than a quarter of unique questions are boolean. 
Another quarter of unique questions (26.57\%) contain the word \textit{``What``},
6.64\% of questions asks \textit{``Who``} and \textit{``Whose``}, and 4.49\% \textit{``Which``},
about 3\% of questions are  \textit{``When``} and \textit{``Where``}. 
Only 5.95\% ask the question \textit{``how``} excluding (\textit{``how many/much``} and \textit{``how old``}). 
Other questions which do not contain any of these question words constitute 16.73\% of unique questions.
There are datasets where more than 20\% of questions are formulated in such a way that the first token is not one of the considered words: 
Quasar-S (98.8 \%),
SearchQA (98.3 \%),
RACE-C (64.1 \%),
ARC (59.2\%),
TriviaQA (49.6 \%),
HotPotQA (42.0 \%),
Quasar-T (40.7 \%),
QASC (29.4 \%),
MSMARCO (26.6 \%),
NaturalQuestions(23.4 \%),
AmazonQA (22.8 \%),
and SQuAD (21.1 \%). 

See Table~\ref{table:questions_persentage} for more detailed information.

\begin{center}

\begin{table*}
\small
\begin{tabular}{p{2cm}|c|ccccccp{0.8cm}p{0.8cm}p{0.8cm}c}
\bf Dataset& \bf \# of Q  
& \bf what & \bf when & \bf where & \bf which & \bf why & \bf how & \bf who/ whose & \bf how old/ much/ many  & \bf boolean & \bf other\\
\hline

AmazonQA&830954&10.2\%&0.4\%&1.2\%&0.4\%&0.6\%&6.8\%&0.1\%&2.3\%&55.3\%&22.8\%\\
AmazonYesNo&80391&0.2\%&0.2\%&0.0\%&0.0\%&0.0\%&0.1\%&0.0\%&0.0\%&88.3\%&11.1\%\\
ARC&7787&6.0\%&3.4\%&0.4\%&28.2\%&0.9\%&1.4\%&0.0\%&0.2\%&0.2\%&59.4\%\\
bAbI&40000&21.7\%&0&36.9\%&0&3.1\%&5.0\%&3.3\%&5.0\%&25.0\%&0\\
BiPaR&14669&30.9\%&1.9\%&8.7\%&1.7\%&15.7\%&6.5\%&26.4\%&3.0\%&1.9\%&3.2\%\\
BoolQ&15942&0.1\%&0.1\%&0.0\%&0&0&0.0\%&0.0\%&0&97.5\%&2.3\%\\
CoQA&116630&29.7\%&4.1\%&6.6\%&1.6\%&2.7\%&4.6\%&14.7\%&5.4\%&20.7\%&9.9\%\\
CosmosQA&35210&54.6\%&0.2\%&1.6\%&0.7\%&34.2\%&5.2\%&1.1\%&0.3\%&1.2\%&0.9\%\\
DREAM&9934&56.3\%&4.7\%&10.0\%&2.9\%&8.5\%&6.5\%&3.6\%&4.1\%&1.1\%&2.2\%\\
DROP&86945&6.5\%&0.6\%&0.5\%&18.2\%&0.1\%&0.7\%&8.1\%&60.4\%&1.7\%&3.2\%\\
DuoRC&100966&33.1\%&1.0\%&8.6\%&1.2\%&2.5\%&3.4\%&39.5\%&2.6\%&1.8\%&6.3\%\\
emrQA&1980621&16.1\%&0.5\%&0.0\%&0&1.4\%&1.0\%&0.0\%&0.2\%&74.0\%&6.8\%\\
HotPotQA&105253&22.6\%&2.6\%&1.9\%&13.5\%&0.0\%&0.6\%&8.6\%&1.1\%&6.9\%&42.0\%\\
IIRC&12143&20.3\%&5.9\%&2.0\%&12.8\%&0.2\%&6.6\%&11.0\%&20.0\%&11.6\%&9.6\%\\
MCScript&13939&13.9\%&5.8\%&9.4\%&0.5\%&11.6\%&13.4\%&12.2\%&3.8\%&28.6\%&0.7\%\\
MCScript2&19821&42.0\%&27.9\%&11.0\%&0.2\%&0.7\%&3.8\%&8.4\%&0.7\%&0.0\%&5.2\%\\
MCTest 160&639&51.3\%&1.4\%&6.9\%&2.2\%&12.1\%&3.9\%&13.3\%&3.9\%&1.6\%&3.4\%\\
MCTest 500&2000&52.1\%&1.7\%&8.5\%&3.2\%&12.0\%&3.5\%&12.6\%&3.9\%&0.8\%&1.7\%\\
MSMARCO&1009035&35.6\%&2.7\%&3.5\%&1.8\%&1.7\%&11.1\%&3.4\%&5.8\%&7.9\%&26.6\%\\
MovieQA&29888&46.3\%&1.2\%&6.8\%&0.9\%&11.0\%&9.4\%&19.3\%&1.4\%&1.9\%&1.8\%\\
MultiRC&7903&36.6\%&2.3\%&3.9\%&4.0\%&7.0\%&6.7\%&14.3\%&4.6\%&7.2\%&13.4\%\\
NarrativeQA&46764&38.3\%&1.6\%&7.5\%&2.2\%&9.8\%&8.3\%&24.4\%&2.2\%&0.1\%&5.6\%\\
NaturalQuestions&315104&15.5\%&13.1\%&10.1\%&2.9\%&1.2\%&2.3\%&25.3\%&3.8\%&2.6\%&23.4\%\\
NewsQA &119632 &44.3\%  &4.1\%  &7.1\%  &2.2\%  &0.1\% &0.9\% &19.8\% &5.9\% &3.9\% &11.7\%  \\
OpenBookQA&5930&9.5\%&2.9\%&1.2\%&10.6\%&0.6\%&0.7\%&0.3\%&0.3\%&0.5\%&73.5\%\\
PubMedQA &1000   &0    &0      &0  &0  &0      &0   &0      &0 &64.1\% &35.9\%\\
QASC&9980&61.8\%&1.5\%&2.1\%&1.6\%&0.5\%&2.3\%&0.2\%&0.2\%&0.4\%&29.4\%\\
QuAC&90922&35.0\%&5.2\%&3.5\%&0.7\%&2.8\%&6.6\%&5.3\%&1.4\%&36.6\%&2.9\%\\
Quasar-S&37362&0.0\%&0.0\%&0.0\%&0&0&0.0\%&0.0\%&0&1.1\%&98.8\%\\
Quasar-T&41102&32.0\%&0.5\%&2.1\%&10.6\%&0.2\%&0.6\%&11.3\%&1.4\%&0.5\%&40.7\%\\
RACE-C&11909&17.7\%&1.1\%&0.3\%&10.0\%&4.4\%&1.4\%&0.6\%&0.3\%&0.1\%&64.1\%\\
RACE&51526&35.6\%&1.8\%&2.3\%&23.1\%&8.5\%&4.3\%&2.6\%&3.0\%&0.4\%&18.4\%\\
ReClor&6138&0.2\%&0&0&56.5\%&0.0\%&0.0\%&0&0&0.0\%&43.2\%\\
SearchQA&163981&0.1\%&0.7\%&0.0\%&0.0\%&0.0\%&0.1\%&0.0\%&0.0\%&0.7\%&98.3\%\\
ShaRC&24160&0&0&0&0&0&0&0&0&85.4\%&14.6\%\\
SQuAD&98160&43.4\%&6.3\%&3.8\%&4.7\%&1.4\%&3.3\%&9.7\%&6.1\%&1.2\%&20.1\%\\
SQuAD2&142183&46.0\%&6.1\%&3.6\%&4.3\%&1.4\%&3.2\%&9.9\%&5.8\%&1.0\%&18.7\%\\
SubjQA&10093&17.8\%&0.7\%&3.2\%&0.8\%&2.0\%&56.4\%&0.4\%&2.2\%&16.5\%&0\\
TriviaQA&800827&18.6\%&0.3\%&0.8\%&19.0\%&0.0\%&0.3\%&9.9\%&1.2\%&0.3\%&49.6\%\\
TurkQA&53700&34.8\%&5.7\%&6.9\%&1.2\%\%&0.2\%&0.9\%&6.9\%&1.7\%&25.7\%&16.0\%\\
TyDi&14378&29.0\%&20.5\%&4.8\%&1.4\%&0.8\%&9.0\%&11.8\%&13.5\%&8.6\%&0.6\%\\
WikiMovies&216453&50.6\%&1.7\%&0.0\%&10.4\%&0.0\%&7.7\%&17.3\%&0&2.2\%&10.1\%\\
WikiQA&1242&54.5\%&9.1\%&8.9\%&0&0&6.5\%&13.5\%&7.5\%&0&0\\
\end{tabular}
\caption{The percentage of question words per dataset.}
\label{table:questions_persentage}
\end{table*}

\end{center}

\end{document}